\documentclass[runningheads]{llncs}

\usepackage{eccv}

\usepackage{eccvabbrv}

\usepackage{amsmath,amsfonts,bm}

\def\eqref#1{equation~\ref{#1}}

\def\1{\bm{1}}

\DeclareMathAlphabet{\mathsfit}{\encodingdefault}{\sfdefault}{m}{sl}
\SetMathAlphabet{\mathsfit}{bold}{\encodingdefault}{\sfdefault}{bx}{n}

\usepackage{url}
\usepackage{cancel}

\usepackage{wrapfig}

\usepackage{epsfig}
\usepackage{graphicx,caption}
\usepackage{amsmath}
\usepackage{amssymb}

\usepackage{booktabs}
\usepackage{adjustbox}
\usepackage{dblfloatfix} 

\usepackage{cuted}
\usepackage{capt-of}

\usepackage{xcolor}         
\newcommand{\wc}[1]{{\color{black}#1}}
\newcommand{\wwc}[1]{{\color{black}#1}}

\usepackage{bm}

\usepackage{multirow}

\usepackage[normalem]{ulem}

\usepackage{url}
\usepackage{wrapfig}
\usepackage{subcaption}
\usepackage{color}
\usepackage{amsmath}
\usepackage{setspace}
\usepackage{multirow}
\usepackage{wrapfig}
\usepackage{bm}
\usepackage{enumitem} 

\usepackage{adjustbox}
\usepackage{changes}

\usepackage{xspace}
\usepackage{comment}
\usepackage[numbers,sort&compress]{natbib}
\usepackage{algorithm} 
\usepackage{mathtools} 
\usepackage{thm-restate} 

\renewcommand{\thepage}{\arabic{page}}

\usepackage{xr-hyper}
\makeatletter

\newcommand*{\addFileDependency}[1]{
\typeout{(#1)}
\@addtofilelist{#1}
\IfFileExists{#1}{}{\typeout{No file #1.}}
}\makeatother

\newcommand*{\myexternaldocument}[1]{%
\externaldocument{#1}%
\addFileDependency{#1.tex}%
\addFileDependency{#1.aux}%
}

\myexternaldocument{supp}

\usepackage[accsupp]{axessibility}  

\newcommand{\indep}{\rotatebox[origin=c]{90}{$\models$}}

\usepackage{hyperref}

\usepackage{orcidlink}

\begin{document}

\title{Enhanced Controllability of Diffusion Models via Feature Disentanglement and Realism-Enhanced Sampling Methods}

\titlerunning{Enhanced Controllability of Diffusion Models via FDiff and GCDM}

\author{Wonwoong Cho\inst{1}\thanks{Performed this work during internship at Adobe Applied Research.}\orcidlink{0000-0003-0898-0341} \and
Hareesh Ravi\inst{2}\orcidlink{0000-0002-3237-1899} \and
Midhun Harikumar\inst{2}\orcidlink{} \and
Vinh Khuc\inst{2}\orcidlink{0009-0001-0368-0350} \and
Krishna Kumar Singh\inst{3}\orcidlink{0000−0002−8066−6835} \and
Jingwan Lu\inst{3}\orcidlink{0000-0002-3598-9918} \and
David Inouye\inst{1}\orcidlink{0000-0003-4493-3358} \and
Ajinkya Kale\inst{2}\orcidlink{0009-0007-1057-8622}}

\authorrunning{Wonwoong Cho et al.}

\institute{Purdue University, West Lafayette IN 47907, USA \and
Adobe Applied Research, San Jose CA 95110, USA \and
Adobe Research, San Jose CA 95110, USA
}

\maketitle

\begin{abstract}
  As Diffusion Models have shown promising performance, a lot of efforts have been made to improve the controllability of Diffusion Models. However, how to train Diffusion Models to have the disentangled latent spaces and how to naturally incorporate the disentangled conditions during the sampling process have been underexplored. In this paper, we present a training framework for feature disentanglement of Diffusion Models (FDiff). We further propose two sampling methods that can boost the realism of our Diffusion Models and also enhance the controllability. Concisely, we train Diffusion Models conditioned on two latent features, a spatial content mask, and a flattened style embedding. We rely on the inductive bias of the denoising process of Diffusion Models to encode pose/layout information in the content feature and semantic/style information in the style feature. Regarding the sampling methods, we first generalize Composable Diffusion Models (GCDM) by breaking the conditional independence assumption to allow for some dependence between conditional inputs, which is shown to be effective in realistic generation in our experiments. Second, we propose timestep-dependent weight scheduling for content and style features to further improve the performance. We also observe better controllability of our proposed methods compared to existing methods in image manipulation and image translation.
  
\end{abstract}

\section{Introduction}
\label{sec:intro}

Diffusion Models~\citep{sohl2015deep,ho2020denoising} have gained a lot of attention due to their impressive performance in image generation~\citep{dhariwal2021diffusion, dalle2, rombach2022high} and likelihood estimation~\citep{nichol2021improved}. Their applications have been widely and deeply explored as well, such as conditional image generation~\citep{mou2023t2i,zhang2023adding}, personalization~\citep{gal2023an, ruiz2023dreambooth}, and image translation~\citep{kwon2022diffusion}. 

Even though numerous research works have been studied in Diffusion Models, it has remained underexplored how to train Diffusion Models to have disentangled latent spaces, and how to sample from the Diffusion Models while incorporating the decomposed conditions naturally in a controllable way. 

The idea of training generative models with multiple external disentangled latent spaces has been widely explored in GANs ~\citep{choi2020stargan, park2020swapping,huang2018multimodal,lee2018diverse, kwon2021diagonal}. A common strategy across such methods is to learn a structure/content feature $z_c$ to capture the underlying structure (e.g., facial shape and pose in face images) and a texture/style feature $z_s$ to capture global semantic information (e.g. gender, color, hairstyle, and eyeglasses). Similar approaches have been tried in Diffusion Models in ~\citep{kwon2022diffusion, preechakul2022diffusion}, but these techniques are limited to learning a single controllable latent space. Recently, several studies~\citep{Tao2023DisDiff,qi2024deadiff} have proposed to have multiple external disentangled spaces of Diffusion Models. However, the image generation performance or the feature disentangling performance are limited.

\begin{wrapfigure}{r}{6cm}
\centering
\includegraphics[width=0.5\textwidth]{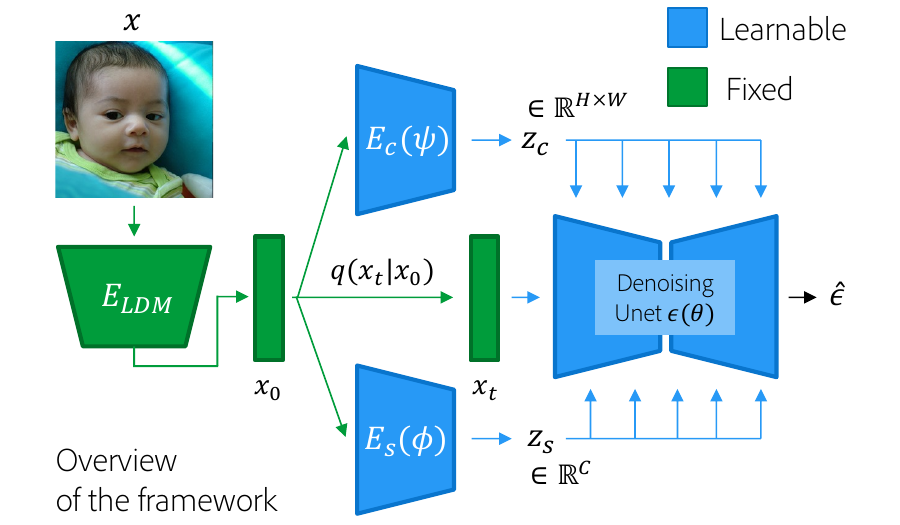}
\caption{Overview of FDiff. Details are provided in Section~\ref{sub:disentangled diffusion models}.}
\label{fig:overview}
\end{wrapfigure}

In this paper, we propose a novel framework FDiff as shown in Fig.~\ref{fig:overview} to effectively learn two latent spaces to enhance the controllability of diffusion models. 
Inspired by ~\citep{park2020swapping, kwon2021diagonal}, we add a \emph{Content Encoder} that learns a spatial layout mask and a \emph{Style Encoder} that outputs a flattened semantic feature to condition the diffusion model during training (Section~\ref{sub:disentangled diffusion models}). The content and style features are injected separately into the UNet~\citep{ronneberger2015u} to ensure that they encode different semantic factors of an image. 




Although decomposing content and style information from an image extends the dimension of the control space, strictly assuming conditional independence between the features during the sampling (i.e., $z_c \, \indep \, z_s | x$) may not always be ideal. For example, \textit{face structure} (e.g. square or round face) encoded in the content feature and \textit{gender} (e.g. male or female) encoded in the style feature \citep{park2020swapping} are not independent given an image. This is because
those two features in reality are correlated in a way that males tend to have square jaws with more protruded chins compared to those of females \citep{Bruce1993}. However, an existing sampling method Composable Diffusion Models (CDM) \citep{liu2022compositional} assumes that conditioning inputs are independent and hence shows unnatural compositions for certain prompts like `a flower' and `a bird' (e.g., Fig.6 in \citep{liu2022compositional} and Fig.~\ref{fig:text2image synthesis} in this paper). 

To this end, we extend the formulation in ~\citep{liu2022compositional} and propose \emph{Generalized Composable Diffusion Models} (GCDM) to support compositions during inference when the conditional inputs are not necessarily independent (Section~\ref{sub:generalized composable diffusion models}). Experiment results consistently show that GCDM has better performance in compositing multiple conditions in terms of realism than CDM by breaking the conditional independence assumption. For instance, from the first row of Fig.~\ref{fig:text2image synthesis}, we can see that the unnatural composition of `a yellow flower' and `a red bird' gets improved by considering the dependence between the conditions and thus GCDM can naturally place the objects together.

 To improve the results further, we leverage the inductive bias~\citep{balaji2022ediffi, choi2021ilvr, choi2022perception} of Diffusion Models that learn low-frequency layout information in the later timesteps and high-frequency or imperceptible details in the earlier timesteps. We use a predefined controllable timestep-dependent weight schedule to compose the content and style codes during generation.

\section{Preliminaries and Related Works}
\label{sec:preliminary knowledge}


Preliminaries and related works on Diffusion Models~\citep{sohl2015deep} such as DDPM~\citep{ho2020denoising} and LDM~\citep{rombach2022high} are provided in 
Section~\ref{sec:preliminaries on diffusion models} in the Supplementary.
\noindent\textbf{Guidance in Diffusion Models:}\\
Some recent works have explored modeling the conditional density $p(x_t|c)$ for controllability. Dhariwal et al.~\citep{dhariwal2021diffusion} proposed to use a pretrained classifier 
but finetuning a classifier that estimates gradients from noisy images, which increases the complexity of the overall process~\citep{ho2022classifier}. Ho et al.~\citep{ho2022classifier} proposed to use an implicit classifier 
while Composable Diffusion Models~\citep{liu2022compositional} (CDM) extend the classifier-free guidance approach to work with multiple conditions assuming conditional independence. 

\noindent\textbf{Conditional Diffusion Models:}\\
Conditional Diffusion Models have been explored in diverse applications showing state-of-the-art performance in text-to-image generation (DALLE2~\citep{dalle2}, Imagen~\citep{imagen}, Parti~\citep{parti}). These methods use pretrained embeddings (e.g., CLIP) that support interpolation but not further editability. Instructpix2pix~\citep{brooks2023instructpix2pix} proposed to generate synthetic paired data via pretrained GPT-3~\citep{brown2020language} and StableDiffusion, with which conditional Diffusion Models are trained. DiffAE~\citep{preechakul2022diffusion} proposed to learn a semantic space that has nice properties making it suitable for image manipulation. However, a single latent space capturing all the information makes it difficult to isolate attributes to manipulate. Recently, ControlNet and T2Iadapter~\citep{mou2023t2i,zhang2023adding} showed impressive performance in conditioning image generation. They use additional auxiliary networks and layers that are trained to encode structure into pretrained text2image Diffusion Models. However, our architecture is particular to reference-based image translation, the proposed GCDM and timestep scheduling are generic and applicable to any multi-conditioned Diffusion Models beyond image translation. 

\noindent\textbf{Disentangling Conditional Diffusion Models:}\\
DisDiff~\citep{Tao2023DisDiff}  is one of the recent closely related works. With pretrained Diffusion Models, they train an additional encoder and a decoder to have disentangled representations. However, their results are limited to low resolution (64) and the disentanglement is less convincing than ours (e.g., Fig. 3 and 4 in their paper).
DEADiff~\citep{qi2024deadiff} is the most recent related work. They propose to train an additional module of pretrained Diffusion Models to extract the content and the style features from the reference image and use them to guide conditional image generation. 
However, as shown in Section~\ref{sec:experiments}, we have observed that their disentangling performance is limited in representing high-level style feature (e.g., facial attributes) and content feature preserving the spatial information.


\noindent\textbf{Inference only Editing in Diffusion Models:}\\
SDEdit~\citep{meng2021sdedit} enables structure-preserving edits while Prompt-to-prompt~\citep{hertz2022prompt} modifies the attention maps from cross-attention layers to add, remove, or reweigh the importance of an object in an image. DiffusionCLIP~\citep{kim2022diffusionclip}, Imagic~\citep{kawar2022imagic} and Unitune~\citep{valevski2022unitune} propose optimization-based techniques for text-based image editing. Textual Inversion~\citep{gal2023an} and DreamBooth~\citep{ruiz2023dreambooth} finetune pretrained models using a few reference images to get personalized models. Though the above techniques are helpful with editing, most of these methods require computationally expensive optimization, modify the weights of pretrained model for each sample, and/or don't support fine-grained controllability for reference-based image translation. The closest related work to ours is DiffuseIT~\citep{kwon2022diffusion}. They enabled reference and text-guided image translation by leveraging Dino-VIT~\citep{caron2021emerging} to encode content and style. However, their approach requires costly optimization during inference and doesn't support controlling the final generation.

\section{Proposed Method}
\label{sec:methods}

\begin{figure*}[t]
\begin{minipage}[c]{0.53\textwidth}
        \includegraphics[width=\textwidth]{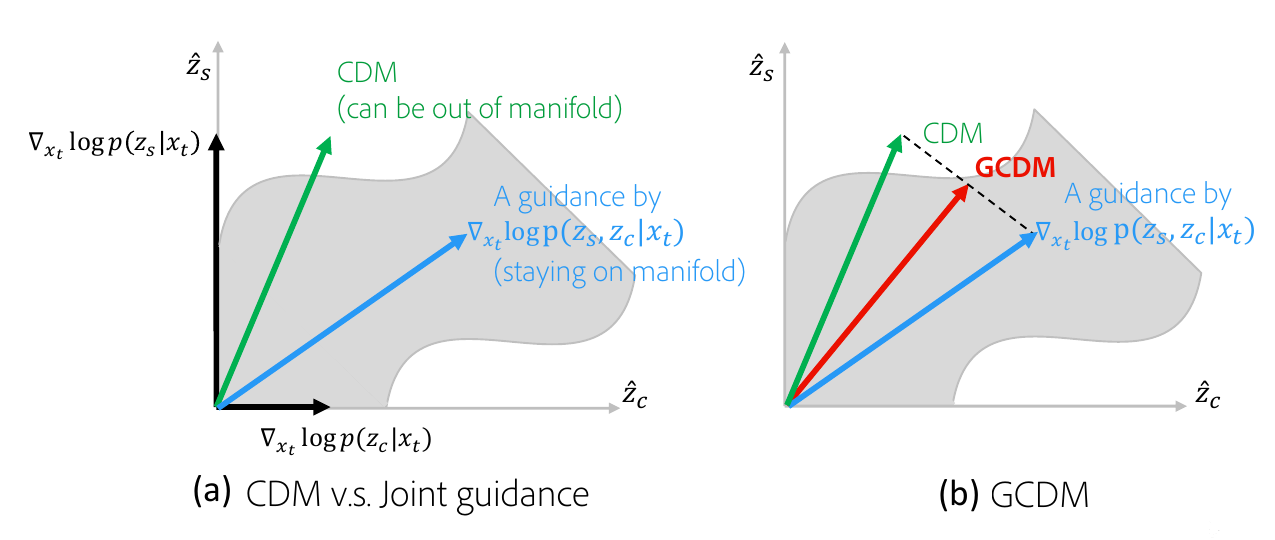}
    \end{minipage}\hfill
    \begin{minipage}[c]{0.45\textwidth}
        \caption{
        Conceptual illustration of CDM and GCDM. (a) The result based on CDM can be outside of manifold while the joint guidance stays on manifold. (b) GCDM trades off between the independent guidance by CDM (stronger conditioning, can be unrealistic) and the joint guidance (more realistic).
        } \label{fig:conceptual_illustration}
    \end{minipage}
\end{figure*}
Our framework is based on the LDM~\citep{rombach2022high} architecture as it is faster to train and sample from, compared to pixel-based diffusion models. Let $x$ be an input image and $E_{LDM}$ and $D_{LDM}$ be the pretrained and fixed encoder and decoder respectively. The actual input space for our diffusion model is the low-dimensional latent space $x_0=E_{LDM}(x)$. The output of the reverse diffusion process is the low dimensional latent $\hat{x}_0$ which is then passed through the pretrained decoder as $\hat{x}=D_{LDM}(\hat{x}_0)$.


\subsection{Learning Content and Style Latent spaces of FDiff}
\label{sub:disentangled diffusion models}


Inspired by DiffAE~\citep{preechakul2022diffusion} and similar approaches in GANs~\citep{kwon2021diagonal}, we introduce a content encoder $E_c(\, \cdot \, ; \psi)$ and a style encoder $E_s(\, \cdot \, ; \phi)$ in our framework as shown in Fig.~\ref{fig:overview}. The objective for training FDiff is formulated as:
\begin{align}
    \min_{\theta,\psi,\phi} \mathop{\mathbb{E}_{x_0,\epsilon_t}} \left[ \lVert \epsilon_t - \epsilon(x_t, t, E_c(x_0 ; \psi), E_s(x_0 ; \phi) ; \theta) \rVert_2^2 \right],
\end{align}
where $x_t$ is from the forward process, i.e., $x_t=q(x_t|x_0)$. To ensure that the encoders capture different semantic factors of an image, we design the shape of $z_s$ and $z_c$ asymmetrically as done in ~\citep{park2020swapping,tumanyan2022splicing,huang2018multimodal,lee2018diverse, kwon2021diagonal,cho2019image}. 
The content encoder $E_c(x_0;\psi)$ outputs a spatial layout mask $z_c \in \mathbb{R}^{1\times \frac{h}{4}\times \frac{w}{4}}$ where $w$ and $h$ are the width and height of $x_0$ latent. On the other hand, $E_s(x_0;\phi)$ outputs $z_s \in \mathbb{R}^{512 \times 1 \times 1}$ after global average pool layer to capture global high-level semantics. At each layer of the denoising UNet $\epsilon(\, \cdot \, ; \theta)$, the style code $z_s$ is applied using channel-wise affine transformation while the content code $z_c$ is applied spatially. In specific, let $\odot$ denote the element-wise product (i.e., Hadamard product), let $\otimes$ denote outer product, and let $\mathbf{1}$ denote all one's vector/matrix where subscript denotes dimensionality. We define a variant of adaptive group normalization layer (AdaGN) in the UNet as:


\begin{align}
\begin{aligned}
\text{AdaGN}(h^\ell)=&[\mathbf{1}_{C} \otimes (\mathbf{1}_{H,W} + t_1\varphi^\ell(z_c))]
\odot [ 
(\mathbf{1}_C+\zeta^\ell(z_s)) \otimes \mathbf{1}_{H,W}]  \\
& \odot [[\bm{t}_2 \otimes \mathbf{1}_{H,W}] \odot \text{GN} (h^\ell) + [\bm{t}_{3} \otimes \mathbf{1}_{H,W}]], 
\label{eq:how_to_combine_content_and_style}
\end{aligned}
\end{align}

where $\varphi^\ell$ is the spatial-wise content-specific term operating down or upsampling at $\ell$-th layer to make the dimensions of $\varphi^\ell(z_c)$ and $h^\ell$ match. $\zeta^\ell$ is the channel-wise style-specific term that is implemented as an MLP layer. $h^\ell$ is (hidden) input to the $\ell$-th layer, and $t$ terms are timestep-specific adjustment terms inspired by Eq. 7 in DiffAE~\citep{preechakul2022diffusion}\footnote{The proposed AdaGN by DiffAE is formulated as $z_s(t_2\text{GN}(h)+t_3)$}. When $C,H,W$ are all 1 (i.e., the scalar case),  the equation can be simplified to reveal the basic structure as $(1 + t_1\varphi(z_c))(1 +\zeta(z_s))(t_2 \text{GN}(h) + t_3)$. $t_2$ and $t_3$ are affine parameters adjusting the input $h$ to the specific timestep. $t_1$ is a content-specific timestep adjustment term allowing the denoising networks $\epsilon(\, \cdot \, ; \theta)$ to decide to strengthen/weaken the content effect since the content feature mostly encodes the low-frequency information, which doesn't play an important role in denoising process over the earlier timesteps~\cite{choi2021ilvr,choi2022perception, balaji2022ediffi}. The $(1+\underline{\hspace{1em}})$ structure reveals the residual architecture of the content and style. It enables the denoising networks $\epsilon(\, \cdot \, ; \theta)$ to work even when the content or/and the style condition are not given.

\subsection{Generalized Composable Diffusion Models (GCDM)}
\label{sub:generalized composable diffusion models}

Once the disentangled representations $z_c$ and $z_s$ are obtained, the next thing we need to consider is how to incorporate them into the Diffusion sampling process. Although CDM has provided a fundamental ground for composing multiple conditions, it has an inherent limitation that conditional independence is assumed (i.e., $z_c \, \indep \, z_s | x$), which may not always hold in practice. 

To this end, we introduce GCDM derived by breaking the conditional independence assumption. The conceptual benefits of GCDM over CDM are illustrated in Fig.~\ref{fig:conceptual_illustration}. (a) shows an example that the content and the style guidances from CDM generate unrealistic samples because the naively combined guidance (in green) could be outside the manifold. On the contrary, the joint guidance helps keep the generation within the manifold. (b) visualizes the proposed GCDM which can be seen as a linear interpolation between CDM and the joint guidance. On top of the limited control space of CDM, GCDM can additionally guide the sampling process more realistic. 

\textbf{Deriving GCDM formulation.} We omit the parameter notation $\theta$ of the denoising networks $\epsilon$ for brevity purposes.

Diffusion Models $\epsilon(x_t,t)$ can be seen as estimating the score of data distribution $\nabla_{x_t} \log p(x_t)$\footnote{It can be also seen as gradient-based MCMC sampling of Energy-Based Models~\cite{liu2022compositional}}~\citep{6795935,song2020denoising,liu2022compositional,song2021scorebased}, from which we can derive the GCDM formulation.  We first start from conditional density $p(x_t | z_c, z_s)$ that we want to have during the sampling process. Taking log and partial derivative w.r.t. $x_t$, we can have

\begin{align}
    \nabla_{x_t} \log p(x_t | z_c,z_s) = \nabla_{x_t} \log p(x_t,z_c,z_s) = \nabla_{x_t} \log p(x_t)p(z_c,z_s|x_t).
\end{align}

As opposed to CDM, we do not assume conditional independence, so that the $p(z_c,z_s|x_t)$ term cannot be easily decomposed into $p(z_c|x_t)p(z_s|x_t)$. By breaking the conditional independence assumption, we have 

\begin{align}
    \nabla_{x_t} \log p(x_t | z_c,z_s)  = \nabla_{x_t} \log p(x_t) p(z_c|x_t) p(z_s|x_t) \left( \frac{p(z_c,z_s|x_t)}{p(z_c|x_t)p(z_s|x_t)} \right).
\end{align}

The terms having a form of $\nabla_{x_t} \log p(z|x_t)$ can be seen as implicit classifiers which can be decomposed into $\nabla_{x_t} \log \frac{p(x_t|z)}{p(x_t)}$, where $p(x_t)$ and $p(x_t|z)$ can be modeled by Diffusion Models. By decomposing the implicit classifiers, we have

\begin{align}
    \nabla_{x_t} \log p(x_t | z_c,z_s) = \nabla_{x_t} \log \frac{p(x_t|z_c) p(x_t|z_s)}{p(x_t)}  \left( \frac{p(x_t|z_c,z_s)p(x_t)}{p(x_t|z_c)p(x_t|z_s)} \right).
\end{align}

Similar to previous studies~\citep{ho2021classifierfree,liu2022compositional},
we can add hyperparameters to control each of the guidance terms and substitute the score function with the denoising networks $\epsilon$, with which we can finally have GCDM formulation in \autoref{def:gcdm}. Please see Section~\ref{sec:derivation} in the Supplementary for the full derivations.

\begin{definition}[Generalized Composable Diffusion Models (GCDM)]
\label{def:gcdm}
The score function of GCDM is the unconditional score function plus a convex combination of joint and independent guidance terms formalized as:
\begin{align}
    \nabla_{x_t} \log \tilde{p}_{\alpha, \lambda, \beta_c, \beta_s}(x_t|z_c,z_s)  \triangleq & \epsilon(x_t,t) + \alpha \Bigl[ \lambda (\underbrace{\epsilon(x_t, t, z_c, z_s) - \epsilon(x_t,t)}_{\nabla_{x_t} \log p(z_c,z_s|x_t)}) \\
     &+ (1-\lambda) \sum_{i=\{c, s\}} \beta_{i} \Bigl( \underbrace{\epsilon(x_t,t,z_i)- \epsilon(x_t,t)}_{\nabla_{x_t} \log p(z_i|x_t)} \Bigr) \Bigr] \,, \nonumber
     \label{eq:definition}
\end{align}
where $\alpha \geq 0$ controls the strength of conditioning,  $\lambda \in [0,1]$ controls the trade-off between joint and independent conditioning, and $\beta_c$ and $\beta_s$ controls the weight for the content and style features respectively under the constraint that $\beta_c+\beta_s= 1$.
%
%
\end{definition}
Intuitively, we can strengthen the style effect by increasing $\beta_s$ and vice versa for the content. We can also increase $\lambda$ if we want to generate more realistic outputs. $\alpha$ can control the overall strength of the guidance terms. Visualizations of the effects of each hyperparameter are provided in Fig.~\ref{fig:FFHQ hyperparameters} and Section~\ref{subsec:params}.

We next show some of the interesting features of GCDM. First, GCDM generalizes simple joint guidance, CDM, and Classifier-Free Guidance~\citep{ho2021classifierfree} (CFG).




\begin{proposition}[GCDM Generalizes Joint Guidance, CDM and CFG]
\label{pro:gcdm}
If $\lambda=1$, then GCDM simplifies to joint guidance:
\begin{align}
    \nabla_{x_t} \log \tilde{p}_{\lambda=1}(x_t|z_c,z_s) &=
    \underbrace{\epsilon(x_t,t) + \alpha (\epsilon(x_t,t,z_c,z_s)- \epsilon(x_t,t))}_{\text{Joint Guidance}} = \nabla_{x_t} \log p(x_t|z_c,z_s)\,.
\end{align}
If $\lambda=0$, then GCDM simplifies to CDM:
\begin{align}
    \nabla_{x_t} \log \tilde{p}_{\lambda=0}(x_t|z_c,z_s) &=
    \underbrace{\epsilon(x_t,t) + \alpha \Bigl[ \sum_{i=\{c, s\}} \beta_{i} (\epsilon(x_t,t,z_i)- \epsilon(x_t,t)) \Bigr]}_{CDM} \, .
\end{align}
If $\lambda=0$ and $\beta_s=0$, then GCDM simplifies to CFG:
\begin{align}
    \nabla_{x_t} \log \tilde{p}_{\lambda=0, \beta_s=0}(x_t|z_c,z_s) &=
    \underbrace{\epsilon(x_t,t) + \alpha \beta_{c} (\epsilon(x_t,t,z_c)- \epsilon(x_t,t))}_{CFG} \,.\label{eq:proposition2}
\end{align}
\end{proposition}
The proof is simple from inspection of the GCDM definition. Second, the GCDM PDF $\tilde{p}$ from \autoref{def:gcdm} is proportional to a nested geometric average of different conditional distributions.
\begin{corollary}
\label{cor:gcdm}
The GCDM distribution $\tilde{p}$ is proportional to  nested geometric averages of conditional distributions of $x_t$:
\begin{align}
    \tilde{p}_{\alpha, \lambda, \beta_c, \beta_s}(x_t|z_c,z_s) \propto
    p(x_t)^{(1-\alpha)} \left[ p(x_t | z_c, z_s) ^ {\lambda}  \Bigl( p(x_t | z_c)^{\beta_c} p(x_t | z_s)^{(1-\beta_c)} \Bigr)^{(1-\lambda)} \right]^{\alpha} \,.
\end{align}
\end{corollary}

The outermost geometric average is between an unconditional and conditional model. Then inside we have the geometric average of the joint and independent conditional, and finally inside the independent conditional we have a geometric average of the independent conditionals. The derivation is provided in Section~\ref{sec:derivation for corollary 3.3} in the Supplementary.





\begin{wrapfigure}{r}{9cm}
\centering
\includegraphics[width=0.75\textwidth]{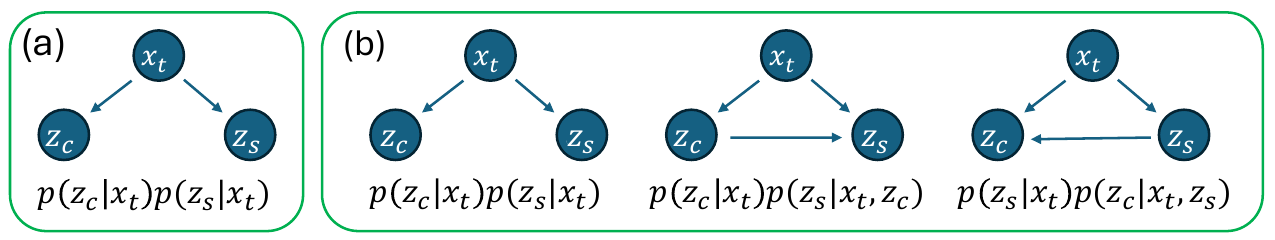}
\caption{Graphical comparisons of (a) CDM and (b) GCDM.}
\label{fig:conceptual difference between CDM and GCDM}
\end{wrapfigure}

\textbf{How can GCDM be more realistic than CDM?} In theory, both GCDM and CDM formulations are derived from the classifier guidance $\nabla_{x_t} \log p(z_c,z_s|x_t)$ as shown in Eq. (7) and (13) in the Supplementary. The better the classifier, the more realistic the guided samples could be. The implicit classifier $p(z_c,z_s|x_t) \propto \frac{p(x_t|z_c,z_s)}{p(x_t)}$ of CDM only considers the conditional independence case (Fig.~\ref{fig:conceptual difference between CDM and GCDM} (a)) while the one in GCDM considers all of the cases of dependency (Fig.~\ref{fig:conceptual difference between CDM and GCDM} (b)).

\noindent Since conditional independence cannot always hold in practice, assuming $\Tilde{p}(z_c,z_s|x_t)$ is a true conditional, the implicit classifier of GCDM can be a better approximation of $\Tilde{p}$, i.e., $\mathcal{D}(\underbrace{p(z_c|x_t)p(z_s|x_t)}_{\text{CDM}}, \Tilde{p}) \geq \mathcal{D}(\underbrace{p(z_c,z_s|x)}_{\text{GCDM}}, \Tilde{p}),$ 
where $\mathcal{D}$ is an arbitrary divergence between two distributions, and equality holds if $z_c \, \indep \, z_s | x$. Thus, generated samples by GCDM can be more realistic.

\subsection{Timestep Scheduling for Conditioning} 
\label{sub:inductive}
It has been observed in~\citep{choi2021ilvr,choi2022perception, balaji2022ediffi} that low-frequency information, i.e., coarse features such as pose and facial shape are learned in the earlier timesteps (e.g., $0 < \text{SNR(t)} < 10^{-2}$) while high-frequency information such as fine-grained features and imperceptible details are encoded in later timesteps (e.g., $10^{0} < \text{SNR(t)} < 10^{4}$) in the reverse diffusion process. 

Inspired by this, we introduce a weight scheduler for $z_c$ and $z_s$ that determines how much the content and the style conditions are applied to the denoising networks. We use the following schedule:
\begin{align} 
w_c(t) &= \frac{1}{1+\exp{( -a(t-b) )}}, \quad w_s(t) = \frac{1}{1+\exp{( -a(-t+b) )}},
\label{eq:timestep scheduling}
\end{align}
where $a$ is a coefficient for determining how many timesteps content and style are jointly provided while $b$ indicates the timestep at which $w_s(t) \ge w_c(t)$. In most experiments, we use $a=0.025$ and $b=550$. When applying, we use weighted form, denoted $\bar{\varphi}$ and $\bar{\zeta}$, of the style and content functions during sampling. They are defined as
$\bar{\varphi}(z_c,t) := w_c(t) \varphi(z_c)$ and $\bar{\zeta}(z_s,t) := w_s(t) \zeta(z_s)$, which respectively replace $\varphi$ and $\zeta$ in Eq.~\ref{eq:how_to_combine_content_and_style}. We also tried a simple linear weighting and a constant schedule but observed that the proposed schedule gave consistently better results (Section~\ref{subsec:various schedules} in the Supplementary). 


\wwc{We additionally report using timestep scheduling during training. It is an interesting future direction showing better decomposition of content and style (Section~\ref{subsec:training timestep} in the Supplementary).} 


\section{Experiments}
\label{sec:experiments}

We comprehensively evaluate the proposed training and sampling methods. Implementation details and experimental setup, such as specific descriptions on the baselines and evaluation measures are provided in Section~\ref{sec:details} in the Supplementary.
\subsection{Comparison with Existing Works}
\label{sub: evaluations of entire frameworks}

\begin{figure*}[t]
\centering
\includegraphics[width=0.8\textwidth]{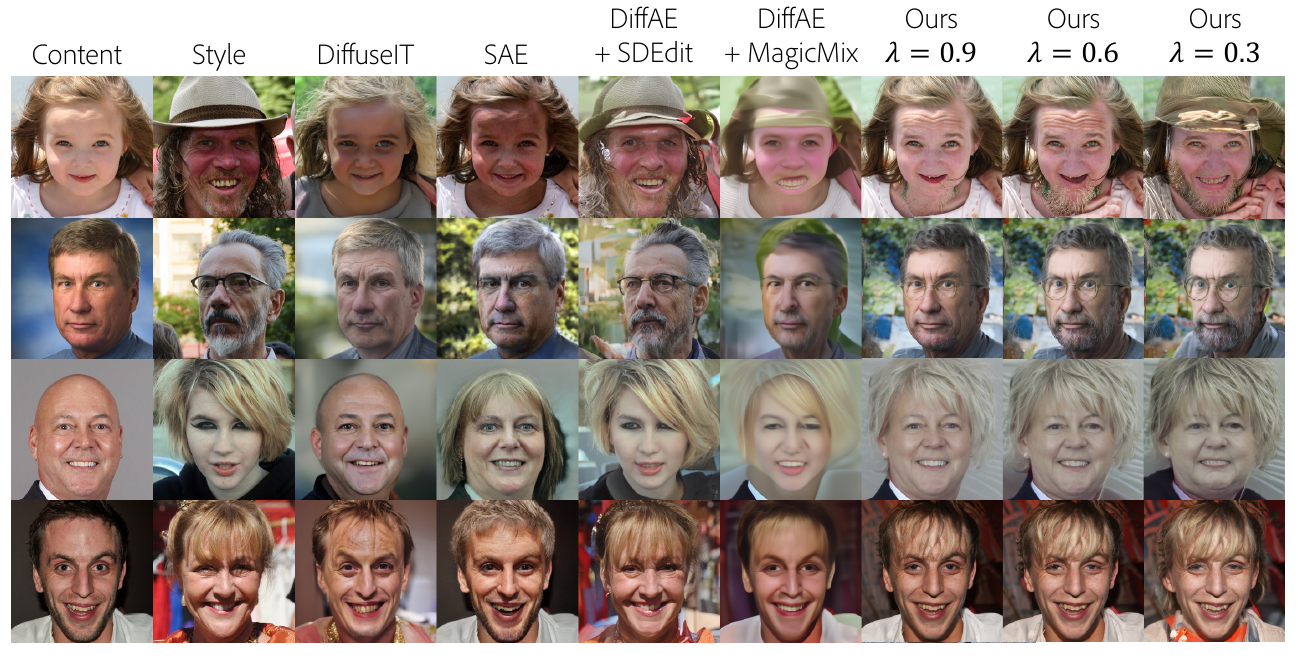}
\caption{Comparison of the proposed model with baselines on FFHQ dataset.}
\label{fig:FFHQ baseline comparisons}
\end{figure*}

\begin{figure*}[h]
\centering
\includegraphics[width=0.8\textwidth]{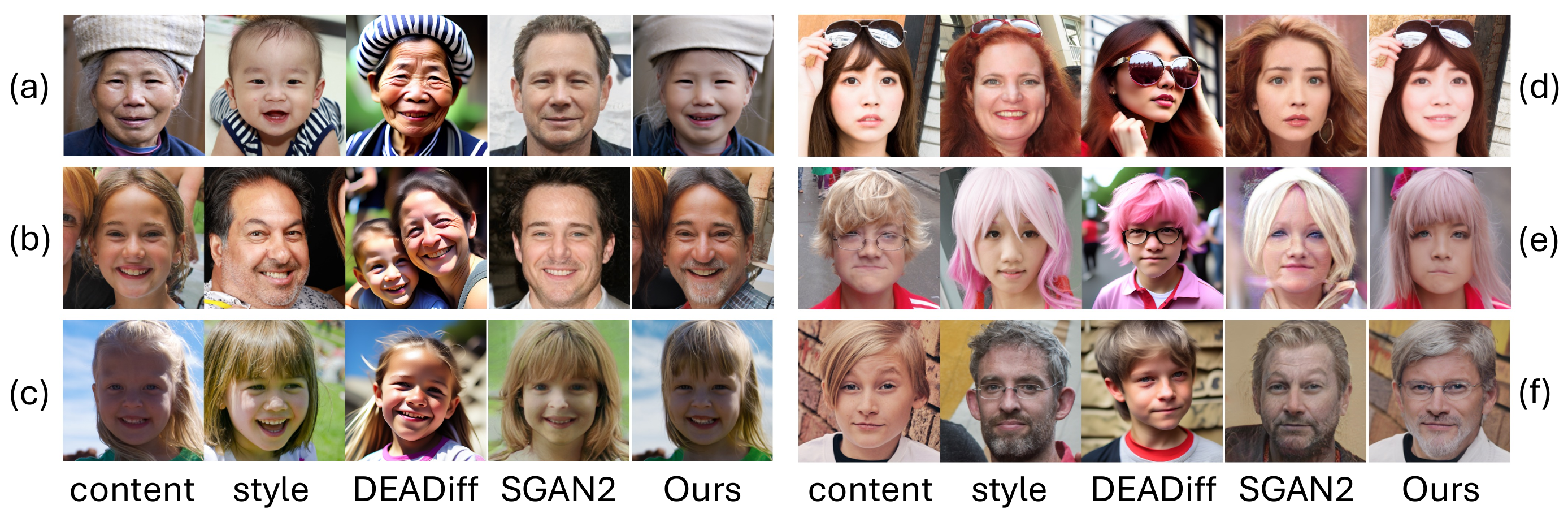}
\caption{Comparison of the proposed model with additional baselines on FFHQ dataset.}
\label{fig:FFHQ additional baseline comparisons rebuttal}
\end{figure*}

\noindent\textbf{Qualitative Results.}\\
Fig.~\ref{fig:FFHQ baseline comparisons} visually shows example generations from different techniques. We observe that DiffAE+SDEdit loses content information while DiffAE+MagicMix generates unnatural images that naively combine the two images. This indicates that a single latent space even with additional techniques such as SDEdit and MagicMix is not suitable for reference-based image translation. Please note that we used the best hyperparameters that they suggested in their papers. 
DiffuseIT and SAE models maintain more content information but do not transfer enough information from the style image and have no control over the amount of information transferred from style. 


One important benefit of our proposed method over the baselines is better controllability. First of all, by manipulating $\lambda$, we can control how much joint guidance is applied. In Fig.~\ref{fig:FFHQ baseline comparisons}, decreasing $\lambda$ indirectly increases the effect of style from the style image when $\beta_c=0$ and $\beta_s=1$, where $\beta_c$ and $\beta_s$ are the weights for each conditional guidance. It is because the smaller $\lambda$ brings more information from the style guidance (ref. $1-\lambda$ term in \autoref{def:gcdm}). For example, the man on the second row has more wrinkles and a beard as $\lambda$ decreases. Second, given a fixed value of $\lambda$, we can control the amount of the content and the style guidance by controlling $\beta_c$ and $\beta_s$ as shown in Fig.~\ref{fig:FFHQ scheduling ablation}. 

Regarding Fig.~\ref{fig:FFHQ additional baseline comparisons rebuttal}, the style representations of DEADiff do not have facial attribute information, and the contents do not preserve detailed spatial information. For example, the color and texture information is transferred from the style while the person's identity is transferred from the content (e.g., (a), (d), and (e)). The results of StarGAN v2 tend to apply the style strongly and to lose the content identity (e.g., (a), (b), and (c)) Due to the inherent limitations of GANs, they also cannot properly generate out-of-mode samples, such as a hat, a baby, and hands (e.g., (a) and (d)).

\begin{table*}[h]
        \captionof{table}{Quantitative comparison using FID and LPIPS on FFHQ dataset.}
\centering
    \begin{adjustbox}{width=0.99\textwidth}
    \begin{tabular}{c c c c c c c c c c}
    \hline 
    & DiffuseIT & SAE & DiffAE + & DiffAE + & StarGAN & DEADiff & Ours & Ours & Ours \\
    & \citep{kwon2022diffusion} & \citep{park2020swapping} & SDEdit~\citep{preechakul2022diffusion,meng2021sdedit} & MagicMix~\citep{preechakul2022diffusion,liew2022magicmix} & v2~\citep{choi2020stargan} & \citep{qi2024deadiff} & ($\lambda=0.9$) & ($\lambda=0.6$) & ($\lambda=0.3$) \\
    \hline
   FID & 29.99 & 25.06 & 26.63 & 84.55 & 48.48 & 76.84 & \textbf{11.99} & 13.40 & 15.45  \\ 
   LPIPS & 0.47 & 0.39 & 0.64 & 0.41 & 0.55 & 0.37 & 0.34 & 0.42 & 0.49 \\
   \hline
    \end{tabular}
    \end{adjustbox}
    \label{table:FFHQ baseline comparisons}
\end{table*}

\noindent\textbf{Quantitative Results.}\\
Table~\ref{table:FFHQ baseline comparisons} shows the quantitative comparison in terms of FID and LPIPS metrics on FFHQ dataset. Please see Section~\ref{sub:exp setup} in the Supplementary for detailed evaluation settings. \wwc{Ours} generate images that are realistic as indicated by the lowest FID scores compared with other models while also performing comparable on diversity as measured by LPIPS. Specifically, DiffAE+SDEdit shows the highest LPIPS but does not show a meaningful translation since it is hard to find any of the content information from the results (e.g., Fig.~\ref{fig:FFHQ baseline comparisons}). StarGAN v2 shows better LPIPS than ours. This is because they tend to apply the style stronger than ours while losing the content information. However, StarGAN v2 show worse FID than ours because their results are limited in representing out-of-mode samples as shown in Fig.~\ref{fig:FFHQ additional baseline comparisons rebuttal} (a) and (d). 

DiffAE+MagicMix shows the worst performance because of its unrealistic generation. SAE and DiffuseIT show lower LPIPS scores than ours, indicating that they transfer relatively little information from the style image onto the generated samples (i.e., less diverse). We can also observe that increasing \wwc{$\lambda$} (when $\beta_c=0$ and $\beta_s=1$) makes LPIPS worse while improving FID. In other words, the stronger \wwc{the joint guidance is the more realistic but less diverse the generated samples are. This verifies our assumption in Fig.~\ref{fig:conceptual_illustration} that the joint component has an effect of pushing the generations into the real manifold.}


\subsection{Effect of GCDM and Timestep Scheduling}

\begin{figure}[t]
    \centering
    \includegraphics[width=\textwidth]{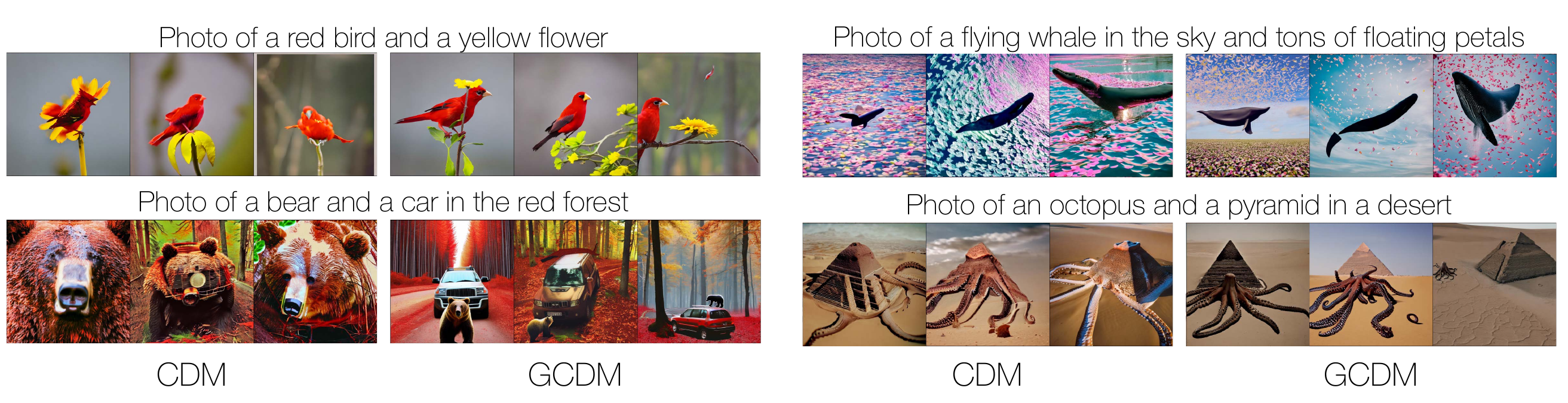}
    \caption{GCDM vs CDM for text-to-image generation with Stable Diffusion. We can observe that CDM generates unnatural images (e.g., blending two objects) that may be out of the real manifold while GCDM ensures realistic generations by breaking the conditional independence assumption (e.g., combining two objects in a realistic way)}
    \label{fig:text2image synthesis}
\end{figure}

\begin{table}[h]
        
    \begin{minipage}[t]{.38\textwidth}
        \caption{FID comparisons between SAE and our model with CDM and GCDM on AFHQ dataset.}
        \begin{adjustbox}{width=\textwidth}
            \begin{tabular}{c c c c c}
                \hline 
                & SAE & CDM & GCDM & GCDM \\
                & & & ($\lambda=0.9$) & ($\lambda=1.0$) \\\hline
               FID & 9.29 & 10.57 & 9.75 & 8.58\\ 
               LPIPS & 0.45 & 0.59 & 0.59 & 0.57\\
               \hline
               \label{table:AFHQ-CDM-joint-GCDM comparisons}
            \end{tabular}
        \end{adjustbox}
    \end{minipage}\hfill
    \begin{minipage}[t]{.58\textwidth}
        \caption{Comparisons between CDM and GCDM in FFHQ. Best method without timestep scheduling is highlighted in bold and with scheduling is highlighted with *.}
        \begin{adjustbox}{width=\textwidth}
            \begin{tabular}{c c c c c c}
                \hline 
                & \multicolumn{2}{c}{w/o schedule} & \multicolumn{3}{c}{w/ schedule} \\
                \cmidrule(lr){2-3}
                \cmidrule(lr){4-6}
                & CDM &  GCDM  & CDM & GCDM ($\beta_c=1$) & GCDM ($\beta_s=1$)   \\ 
               FID & 21.43 & \textbf{14.46} & 10.50 & 10.21* & 10.61 \\ 
               LPIPS & 0.47 & \textbf{0.51} & 0.31 & 0.28 & 0.33* \\
               \hline
               \label{table:FFHQ with/without timestep scheduling}
            \end{tabular}
        \end{adjustbox}
    \end{minipage}
    
\end{table}
\wwc{We compare SAE~\citep{park2020swapping} (the best-performing baseline) and ours with CDM and GCDM on AFHQ dataset in Table~\ref{table:AFHQ-CDM-joint-GCDM comparisons}. The joint guidance ($\lambda=1$) gets the lowest FID indicating that the generations are more realistic as it pulls the guided results to be within the real data manifold. This can be also understood as a positive effect of GCDM that breaks the conditional independence assumption of CDM. It worths noting that GCDM can be thought of as interpolating between CDM and the joint guidance since FID for GCDM ($\lambda=0.9$) is in between the joint and CDM. By comparing LPIPS and FID of the variants of GCDM, we can see that the outputs become less diverse as realism is increased. SAE
shows worse performance than ours in terms of both diversity and realism. The qualitative comparisons can be found in Fig.~\ref{fig:AFHQ cdm gcdm comparisons} in the Supplementary.} \\
\noindent\textbf{Generalizability of GCDM.}\\
Since GCDM is a general sampling method that can be applied without the disentangled features, we also compare the performance of CDM and GCDM in composing text prompts for text-to-image generation using Stable Diffusion V2\citep{rombach2022high} in Fig.~\ref{fig:text2image synthesis}. The phrases before and after `and' are used as the first and the second conditions (which are the text embeddings in this case). The full sentence is used to represent joint conditioning. 
\begin{figure}[t]
    \begin{minipage}[t]{0.48\textwidth}
        \centering
        \captionsetup{width=.95\linewidth}
        \includegraphics[width=\textwidth]{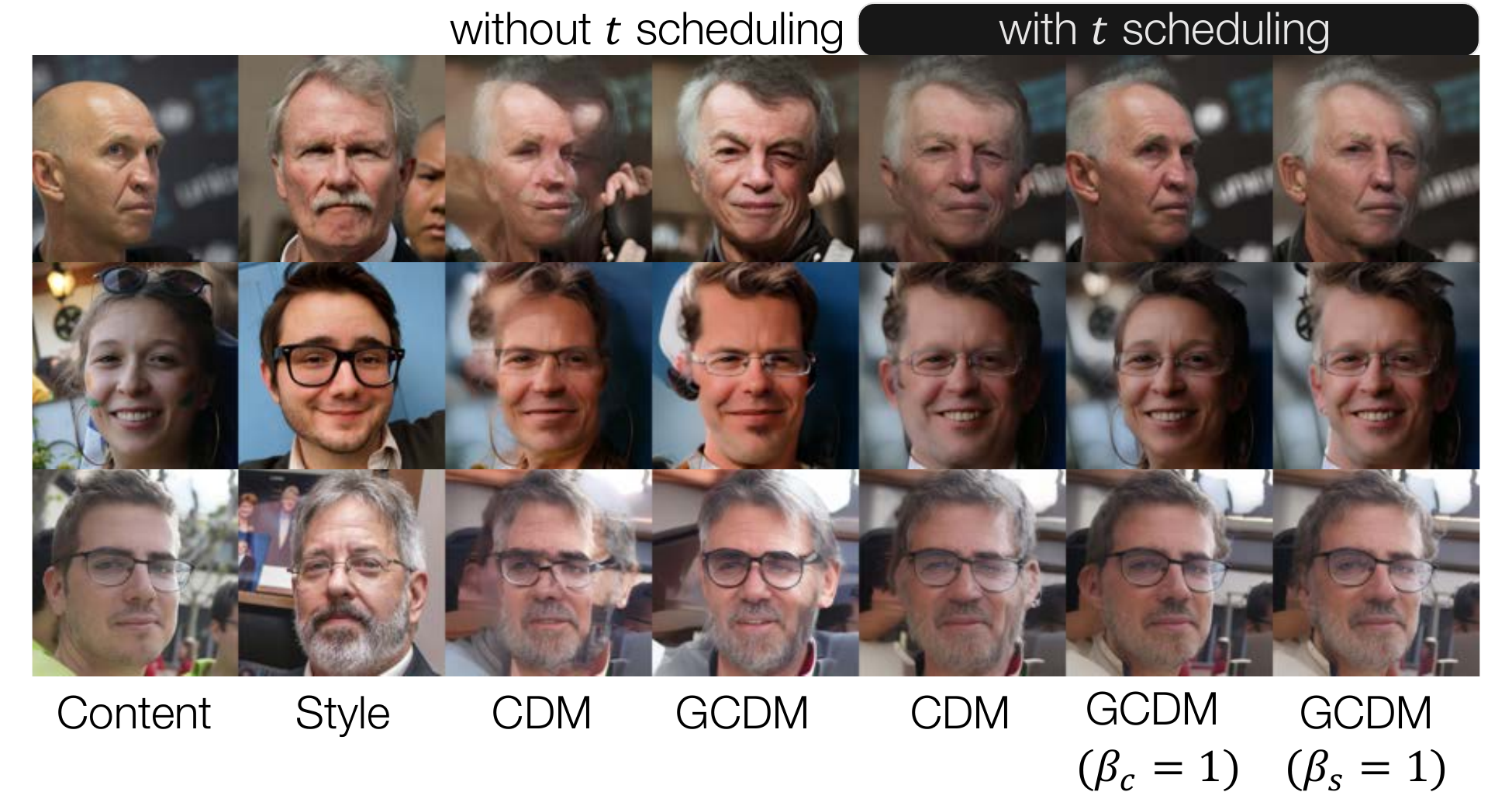}
        \caption{Timestep scheduling improved the results of both CDM and GCDM.}
        \label{fig:FFHQ scheduling ablation}
  \end{minipage}
  \begin{minipage}[t]{0.48\textwidth}
        \centering
        \captionsetup{width=.95\linewidth}
        \includegraphics[width=\textwidth]{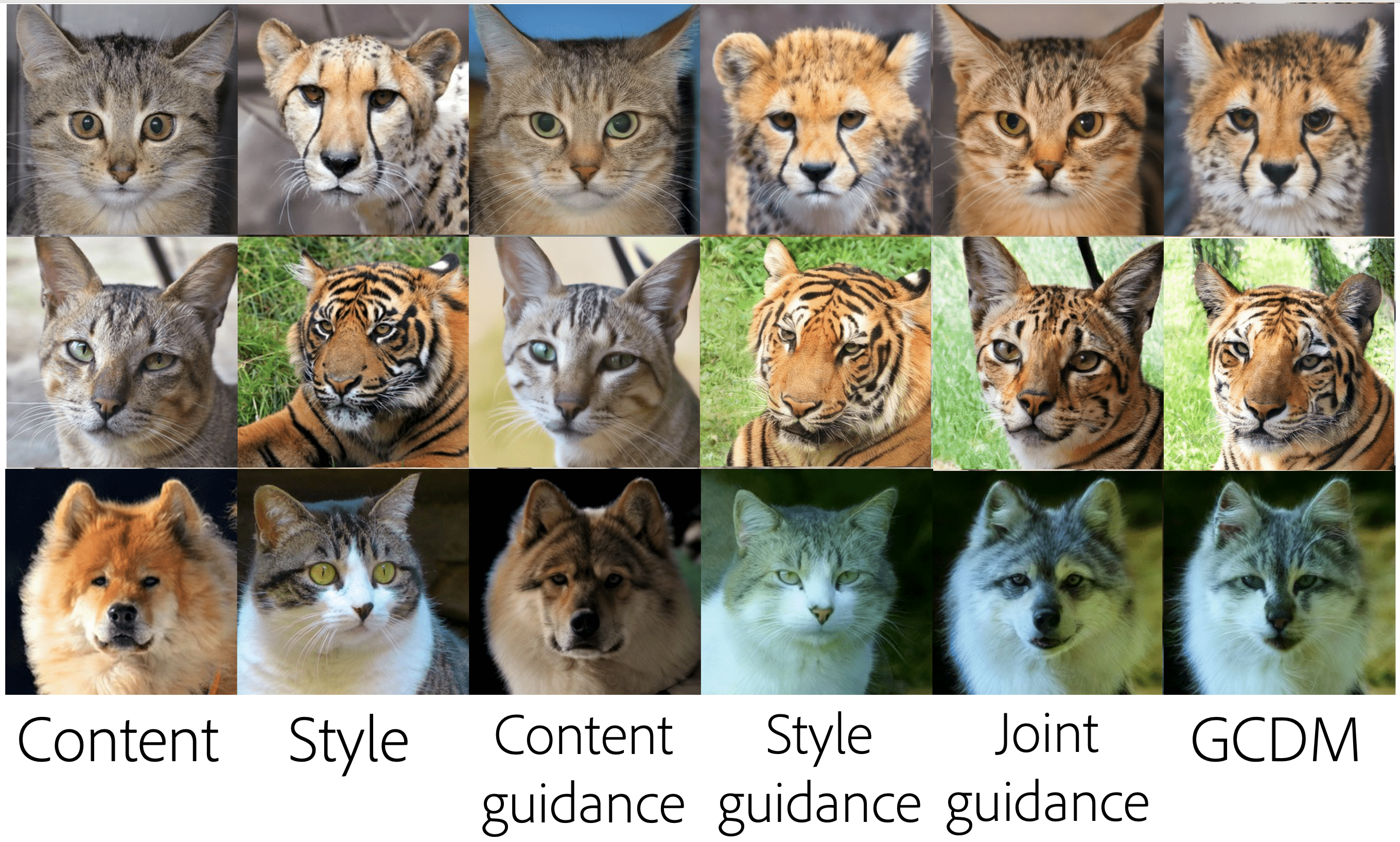}
        \caption{Visualization of the effect of each guidance term on generation.}
        \label{fig:AFHQ visualization of each guidance component}
        
    \end{minipage}%
\end{figure}
As shown in Fig.~\ref{fig:text2image synthesis}, CDM sometimes fails in composing multiple conditions if both conditions contain object information. For example, \emph{the red bird} and \emph{the yellow flower} are unnaturally overlapped in most cases. On the other hand, GCDM consistently shows better compositions in the generated images. This demonstrates that GCDM, a generalized formulation of CDM can show better performance for composing multiple conditions realistically. This is because  GCDM additionally considers the joint guidance by breaking the conditional independence assumption of CDM.  
\\
\noindent\textbf{Effect of Timestep Scheduling.}\\
\wwc{To more carefully analyze the effect of timestep scheduling when combined with GCDM or CDM, we alter the timestep scheduling so that there is at least a 0.1 weight on style or content. Specifically, we change the upper and lower bounds of the sigmoid to be 0.1 and 0.9 in Eq.~\ref{eq:timestep scheduling}, e.g., $w_c'(t) = 0.8w_c(t)+0.1$. The results can be seen in Table~\ref{table:FFHQ with/without timestep scheduling} and Fig.~\ref{fig:FFHQ scheduling ablation}. Without timestep scheduling, GCDM shows better performance in both FID (realism) and LPIPS (diversity) than CDM. 
Combined with timestep scheduling, both CDM and GCDM show meaningful improvements in FID in exchange for losing diversity. This is because timestep scheduling improves content identity preservation. Additionally, timestep scheduling with GCDM variants shows better FID or LPIPS than CDM depending on the strength of guidance terms.}.

\subsection{Choice of Hyperparameters}
\label{subsec:params}
\begin{figure*}[t]
\centering
\includegraphics[width=0.8\textwidth]{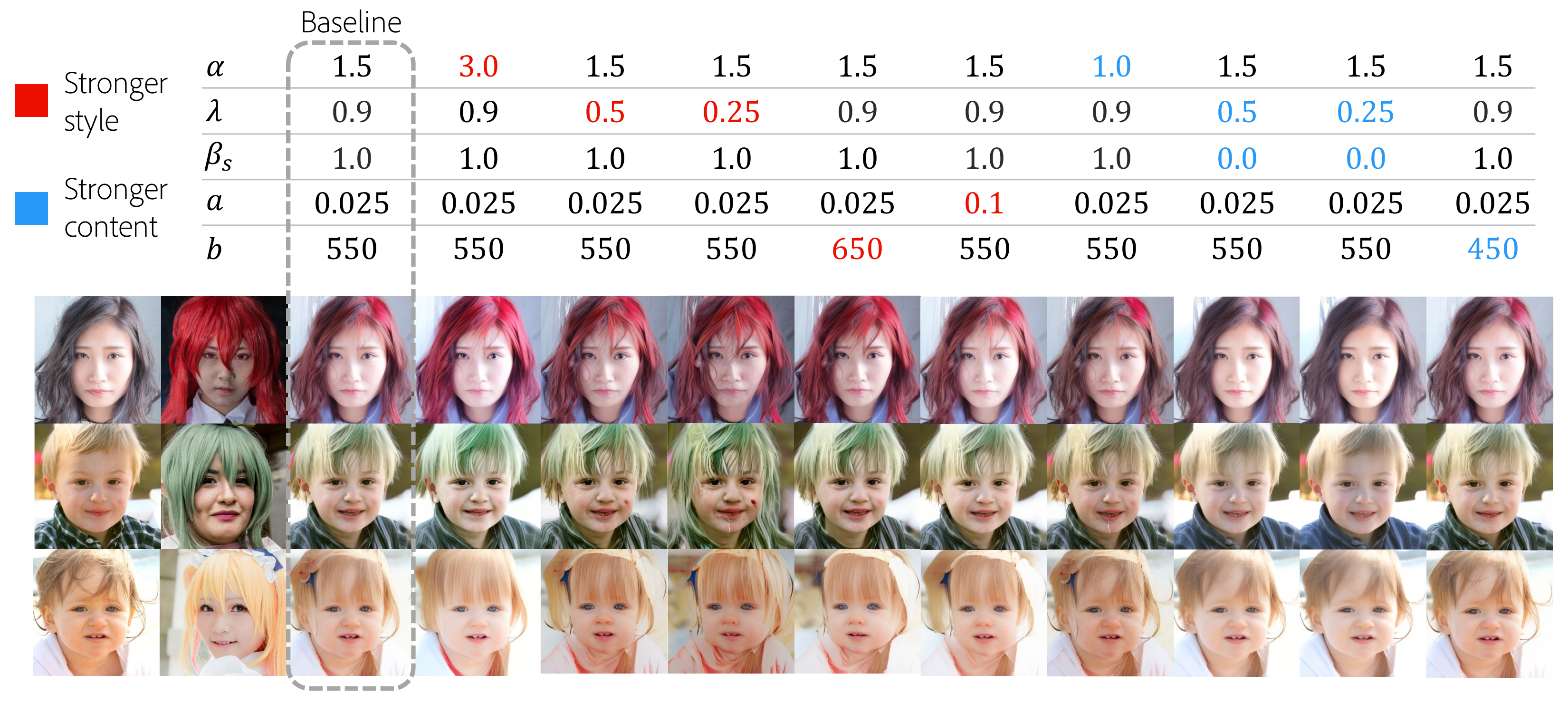}
\caption{Visual guidance on how to set the hyperparameters.}
\label{fig:FFHQ hyperparameters}
\end{figure*}

In Fig.~\ref{fig:FFHQ hyperparameters}, we show the results with various sets of hyperparameters that can be used to control the effect of the content, style, and joint guidance during the sampling. The gray dotted box represents a baseline that we start from, and the rest of the columns show the effects of each hyperparameter. As can be seen in the figure, each hyperparameter can be modified to get desirable results.
Please note that hyperparameter searching is \emph{not mandatory} since the general performance under the baseline setting is good in terms of FID and LPIPS. If a user wants to control more subtly though, here is a tip on how to set the hyperparameters. Increasing $\alpha$ and decreasing $\lambda$ yields stronger style effects. By increasing $\alpha$, since it does not affect the direction of the guidance (but affects the norm), we can expect that the visual features already incorporated are strengthened, such as hair color. Decreasing $\lambda$ though affects the direction by adjusting the guidance components to be incorporated (e.g., Fig.~\ref{fig:conceptual_illustration} (b)), and thus we can expect that new visual features from the style image are applied, such as hair length. However, since a low $\lambda$ indicates the joint guidance is used less, it can pull the generation outside of the manifold so the output can be less realistic. To simplify the control space, we recommend to fix $a$ and $b$ to be 0.025 and 550, respectively.

\subsection{Analysis and Discussion}
\label{sub: analysis}
Further analysis and results on latent interpolation and KNN are in Section~\ref{subsec:interpolation} and \ref{subsec:knn} in the Supplementary. \\ 
\noindent\textbf{Visualization of Each Guidance Term.}\\
The proposed GCDM has guidance from three terms, the joint conditioning and style and content conditionings separately. Fig.~\ref{fig:AFHQ visualization of each guidance component} shows a comparison of the effect of these terms. From the content guidance (column 3), it can be seen that the generated animals are not exactly the same as the content image but have the exact same structure and pose. Similarly, when only style guidance is used (column 4), the pose is random while the style such as color and fur corresponds to the style image. From columns 5-6, it can be observed that the GCDM results have more semantic information from the style guidance than the results by simply using the joint guidance. 

Note that the joint guidance is obtained by breaking the conditional independence assumption of CDM. Since GCDM considers the joint probability $p(z_s,z_c|x)$ as well, it can naturally incorporate both the content and the style information (as argued in Fig.~\ref{fig:conceptual_illustration} (b)). This can be also visual evidence of why GCDM shows the better composition of multiple objects than CDM in Fig.~\ref{fig:text2image synthesis}.\\
\begin{wraptable}{r}{7cm}
    \centering
    \caption{Classifier-based comparisons in FFHQ.}
    \begin{adjustbox}{width=0.55\textwidth}    
    \begin{tabular}{c c c c c c c}
    \hline 
     \multirow{2}{*}{\shortstack{Probability \\ Att. is Equal (\%)}} & \multicolumn{3}{c}{$x_c$} & \multicolumn{3}{c}{$x_s$}\\
     \cmidrule(lr){2-4}
     \cmidrule(lr){5-7}
     
     & Gender & Age & Race & Gender & Age & Race \\
   SAE & 65.95 & 62.36 & 50.40 & 34.05 & 26.40 & 27.91 \\
   Ours ($\lambda=0.9$) & 65.14 & 53.79 & 53.31 & 34.86 & 31.60 & 28.51 \\
   Ours ($\lambda=0.25$) & 26.61 & 25.94 & 31.73 & 73.39 & 56.77 & 44.48 \\
   \hline
    \end{tabular}
    \end{adjustbox}
    
    \label{table:FFHQ classification accuracy comparisons}
\end{wraptable}%

\noindent\textbf{Classifier-based comparisons.} \\
To further understand what kind of attributes are encoded in style and content latent spaces, we use pretrained classifiers to predict the attributes of translated images and compare them with the original style and content images. We sample 2000 random images from the test set to use as $x_c$ and another 2000 as $x_s$ to form 2000 content-style pairs. Next, we acquire the translated output $x_o$ and corresponding pseudo labels $y_c$, $y_s$, and $y_o$ by leveraging an off-the-shelf pretrained attribute classifier (EasyFace). 
In Table~\ref{table:FFHQ classification accuracy comparisons}, we show the probabilities that the final generated image $x_o$ has an attribute from content image as $p(y_c^{att} = y_o^{att})$ and likewise for style image. Both ours and SAE are designed to make $z_s$ encode global high-level semantics, e.g., Gender, Age, etc. Thus, methods would show ideal performance if $y_o^{att}=y_s^{att}\neq y_c^{att}$. We see that most global attributes come from the content image for SAE indicating conservative translations from the style image (as seen in Fig.~\ref{fig:FFHQ baseline comparisons} and lower LPIPS in Table~\ref{table:FFHQ baseline comparisons}). In contrast, ours has a controllable way of deciding the strength of attributes from the style image through $\lambda$. Since $\beta_s=1$ is used, the lower $\lambda$ yields the stronger style effects (due to $(1-\lambda)$ term in ~\autoref{def:gcdm}).\\






\begin{figure*}[t]
\includegraphics[width=\textwidth]{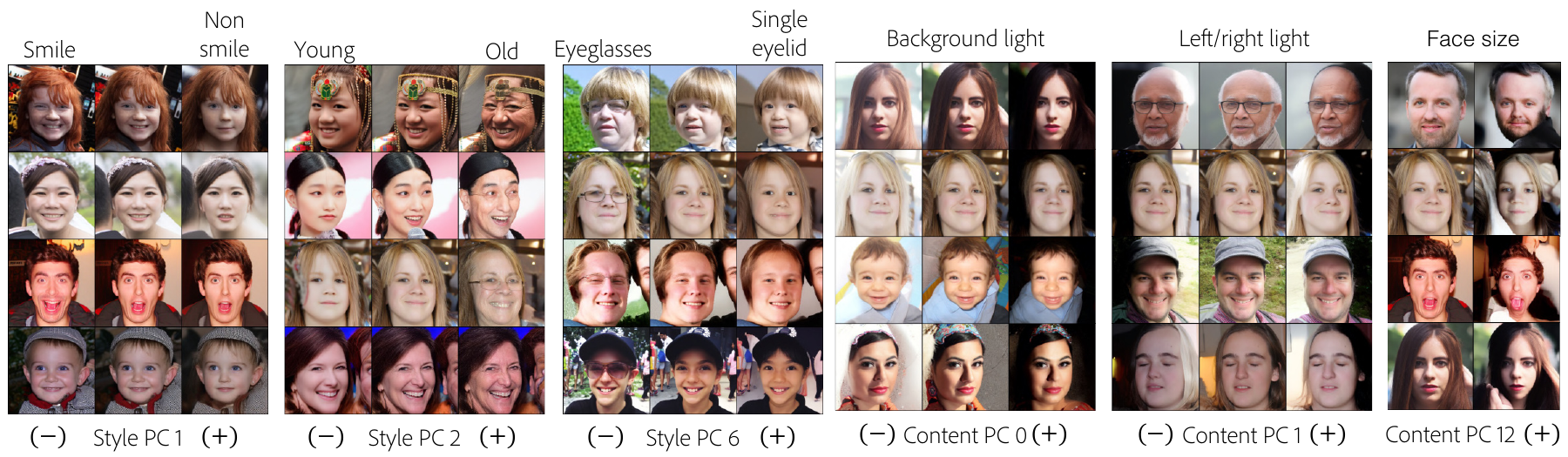}
\caption{Visualizations of Principal Components of the style and the content spaces.}
\label{fig:FFHQ combined_PCA_figure}
\end{figure*}

\noindent\textbf{Visualizing PCA.} \\
To further visualize the encoded information in the content and the style spaces, we apply PCA algorithm on our pretrained latent spaces, inspired by~\citep{ganspace}. Specifically, we first get the style feature and the content feature of all the training images (60000). We then apply PCA to get top 30 Eigenvectors $\mathbf{V_{style}} = \{v_{style}^0, ..., v_{style}^{29}\}$ and $\mathbf{V_{content}} = \{v_{content}^0, ..., v_{content}^{29}\}$. The obtained basis vectors are used for shifting each sample. The results are shown in Fig.~\ref{fig:FFHQ combined_PCA_figure}. We can see that the style space mainly encodes high-level semantic information while the content space mainly encodes structural information as we intended.

\section{Conclusion}
\label{sec:conclusion}
\vspace{-0.1in}
We propose a novel framework FDiff for disentangling the latent space of Diffusion Models. Our content and style encoders trained along with the Diffusion Models and do not require additional objectives or labels to learn to decompose style and content from images. We further propose two sampling methods. The proposed GCDM extends CDM for a more generalized scenario by breaking the conditional independence assumption. It shows better performance when compared with CDM for translation as well as compositing text prompts. We also show that timestep-dependent weight schedules for conditioning inputs can help improve overall results and controllability. \wwc{Additionally, it is shown that the learned latent spaces have desirable properties from PCA-based and interpolation-based experiments. Quantitative and qualitative evaluation shows the benefits of the proposed sampling techniques.}

\noindent\textbf{Acknowledgments.} W.C. started this work while on an internship with Adobe. W.C. and D.I. acknowledge partial support from ARL (W911NF-2020-221). Any opinions, findings, and conclusions or recommendations expressed in this material are those of the authors and do not necessarily reflect the views of the sponsor.

\clearpage  

%
%
\bibliographystyle{splncs04}
\bibliography{main}

\clearpage

\appendix

\section{Overview}
Our contributions can be summarized as follows:

\begin{itemize}
    \item \noindent\textbf{Content and Style Latent Space (FDiff):}\\
We propose a Diffusion training framework FDiff for obtaining the disentangled content and style latent spaces that correspond to different semantic factors of an image. This lets us control these factors separately to perform reference-based image translation as well as controllable generation and image manipulation.
    \item \noindent\textbf{Generalized Composable Diffusion Model (GCDM):}\\
We extend Composable Diffusion Models (CDM) by breaking the conditional independence assumption to allow for dependency between conditioning inputs. This results in better generations in terms of realism and extended controllability. 
    \item \noindent\textbf{Timestep Scheduling:}\\
We leverage the inductive bias of diffusion models and propose timestep-dependent weight schedules to compose information from content and style latent codes for better translation. 
\end{itemize}


Here is a list of contents in Supplementary material. 
\begin{enumerate}
    \item We provide a full derivation of GCDM formulation in Section~\ref{sec:derivation}. 
    \item We next show a derivation that the GCDM PDF $\tilde{p}$ is proportional to a nested geometric average of different conditional distributions in Section~\ref{sec:derivation for corollary 3.3}.
    \item Preliminaries on Diffusion Models are provided in Section~\ref{sec:preliminaries on diffusion models}.
    \item We explain the training details in Section~\ref{sec:details}.
    \item We visualize the learned style and content space in Section~\ref{sec:controls}; PCA results are provided in the main paper. Here, we additionally provide latent interpolation and KNN results.
    \item We also show additional experiment results on timestep scheduling strategy in Section~\ref{sec:timestep}.
    \item We finally provide some additional results such as text2image synthesis and reference-based image translation in Section~\ref{sec:additional}.
\end{enumerate}

\newpage

\section{Derivation for Definition 1}
\label{sec:derivation}

\subsection{Classifier-free Guidance~\citep{ho2022classifier}}
Assuming we have a single condition $c$, CFG formulation can be derived by:
\begin{align}
    \nabla_{x_t} \log p(x_t|c) & = \nabla_{x_t} \log p(x_t,c) \\
    \nabla_{x_t} \log p(x_t,c) & = \nabla_{x_t} \log p(x_t)p(c|x_t) \\
    & = \nabla_{x_t} \log p(x_t)\frac{p(x_t|c)}{p(x_t)} \\
    & = \nabla_{x_t} \log p(x_t) + \left( \nabla_{x_t} \log p(x_t|c) - \nabla_{x_t} \log p(x_t) \right) \\
    & = \epsilon(x_t, t) + \left( \epsilon(x_t, t, c) - \epsilon(x_t, t) \right).
\end{align}

\noindent In practice, $\epsilon(x_t, t) + \alpha \left( \epsilon(x_t, t, c) - \epsilon(x_t, t) \right)$ is used where $\alpha$ is a temperature controlling the condition effect. Please note that we use $c$ to represent a given single condition.

\subsection{Composable Diffusion Models~\citep{liu2022compositional}}
\begin{align}
    \nabla_{x_t} \log p(x_t|z_c, z_s) &= \nabla_{x_t} \log p(x_t,z_c, z_s) \\
    \nabla_{x_t} \log p(x_t, z_c, z_s) &= \nabla_{x_t} \log p(x_t)p(z_c, z_s|x_t), \quad \text{assuming} \,\,\, z_c \, \indep \, z_s | x \\
    &= \nabla_{x_t} \log p(x_t)p(z_c|x_t)p(z_s|x_t) \\
    &= \nabla_{x_t} \log p(x_t)\frac{p(x_t|z_c)}{p(x_t)}\frac{p(x_t|z_s)}{p(x_t)} \\
    & = \nabla_{x_t} \log p(x_t) + \sum_{i=\{c,s\}} \left( \nabla_{x_t} \log p(x_t|z_i) - \nabla_{x_t} \log p(x_t) \right) \\
    & = \epsilon(x_t, t) + \sum_{i=\{c,s\}} \left( \epsilon(x_t, t, z_i) - \epsilon(x_t, t) \right)
\end{align}

Similar to the Classifier-free Guidance, hyperparameters for controlling the weight of each condition are used, i.e., $\epsilon(x_t, t) + \sum_{i=\{c,s\}} \alpha_i \left( \epsilon(x_t, t, z_i) - \epsilon(x_t, t) \right)$.

Now we introduce how to derive the components of GCDM formulation.

\newpage

\subsection{Generalized Composable Diffusion Models}
For brevity purposes, we omit the term that is canceled out because it is constant w.r.t. $x_t$, e.g., $\nabla_{x_t} \log p(z_c,z_s) = 0$ and  $\nabla_{x_t} \log p(z_c) = 0$.

\begin{align}
    \nabla_{x_t} \log p(x_t|z_c,z_s) &= \nabla_{x_t} \log p(x_t,z_c,z_s) \\
    \nabla_{x_t} \log p(x_t,z_c,z_s) &= \nabla_{x_t} \log p(x_t)p(z_c,z_s|x_t), \text{NOT assuming} \,\,\, z_c \, \indep \, z_s | x \\
    &= \nabla_{x_t} \log p(x_t)p(z_c|z_s,x_t)p(z_s|x_t) \\
    &= \nabla_{x_t} \log p(x_t) p(z_s|x_t) \left( \frac{p(z_s|z_c,x_t)p(z_c|x_t)}{p(z_s|x_t)} \right)  \\
    &= \nabla_{x_t} \log p(x_t) p(z_s|x_t) p(z_c|x_t) \left( \frac{p(z_s|z_c,x_t)}{p(z_s|x_t)} \right)  \\
    &= \nabla_{x_t} \log p(x_t) p(z_s|x_t) p(z_c|x_t) \left( \frac{p(z_c,z_s|x_t)}{p(z_c|x_t)p(z_s|x_t)} \right)  \\
    &= \nabla_{x_t} \log \frac{p(x_t|z_s) p(x_t|z_c)}{p(x_t)}  \left( \frac{\frac{p(x_t|z_c,z_s)}{p(x_t)}}{\frac{p(x_t|z_c)p(x_t|z_s)}{p(x_t)^2}} \right) \\
    &= \nabla_{x_t} \log \frac{p(x_t|z_s) p(x_t|z_c)}{p(x_t)}  \left( \frac{p(x_t|z_c,z_s)p(x_t)}{p(x_t|z_c)p(x_t|z_s)} \right)
\end{align}
\begin{align}
    &= - \nabla_{x_t} \log p(x_t) + \nabla_{x_t} \log p(x_t|z_s) + \nabla_{x_t} \log p(x_t|z_c)  \\
    & \quad + \nabla_{x_t} \log p(x_t|z_c,z_s) + \nabla_{x_t} \log p(x_t) - \left( \nabla_{x_t} \log p(x_t|z_c) + \nabla_{x_t} \log p(x_t|z_s) \right) \nonumber \\
    &= -\epsilon(x_t, t) + \epsilon(x_t, t, z_s) + \epsilon(x_t, t, z_c) \nonumber \\
    & \quad + \epsilon(x_t, t, z_c, z_s) + \epsilon(x_t, t) - \left( \epsilon(x_t, t, z_c) + \epsilon(x_t, t, z_s) \right)
\end{align}

By rearranging the terms in Eq. (21) and adding hyperparameters $\alpha$, $\lambda$ and $\{\beta_c, \beta_s\}$, the proposed GCDM method in Definition 1 in the main paper can be obtained. 



\paragraph{\textbf{Clarification of Eq. (28) and Eq. (29).}} 

By Bayes Theorem, Eq. (28) becomes
\begin{align*}
    & \nabla_{x_t} \log \left[ \underbrace{ p(x_t) \frac{p(x_t|z_s)p(z_s)}{p(x_t)} \frac{p(x_t|z_c)p(z_c)}{p(x_t)} }_{\textcircled{\raisebox{-0.9pt}{1}}}  \underbrace{\left( \frac{\frac{p(x_t|z_c,z_s)p(z_c,z_s)}{p(x_t)}}{\frac{p(x_t|z_c)p(z_c)p(x_t|z_s)p(z_s)}{p(x_t)^2}} \right)}_{\textcircled{\raisebox{-0.9pt}{2}}} \right].
\end{align*}

By rearranging $\textcircled{\raisebox{-0.9pt}{1}}$ and $\textcircled{\raisebox{-0.9pt}{2}}$ separately, the above equation becomes

\begin{align*}
    &= \nabla_{x_t} \log \left[ \underbrace{ p(z_s)p(z_c)\frac{p(x_t|z_s)p(x_t|z_c)}{p(x_t)} }_{\text{rearranged from}\, \textcircled{\raisebox{-0.9pt}{1}}} \underbrace{ \left( \frac{\frac{p(x_t|z_c,z_s)}{p(x_t)}}{\frac{p(x_t|z_c)p(x_t|z_s)}{p(x_t)^2}} \right) \left( \frac{p(z_c,z_s)}{p(z_c)p(z_s)} \right) }_{\text{rearranged from}\, \textcircled{\raisebox{-0.9pt}{2}}} \right].
\end{align*}

By canceling out $p(z_c)p(z_s)$ in the first and the last term and by rearranging the equation, we can obtain Eq. (29), i.e.,
\begin{align*}
    &= \nabla_{x_t} \log \left[ \cancel{p(z_s)p(z_c)}\frac{p(x_t|z_s)p(x_t|z_c)}{p(x_t)} \left( \frac{\frac{p(x_t|z_c,z_s)}{p(x_t)}}{\frac{p(x_t|z_c)p(x_t|z_s)}{p(x_t)^2}} \right) \left( \frac{p(z_c,z_s)}{\cancel{p(z_c)p(z_s)}} \right) \right] \\
    &= \nabla_{x_t} \log \left[ \frac{p(x_t|z_s)p(x_t|z_c)}{p(x_t)} \left( \frac{\frac{p(x_t|z_c,z_s)}{p(x_t)}}{\frac{p(x_t|z_c)p(x_t|z_s)}{p(x_t)^2}} \right)  p(z_c,z_s) \right] \\
    &= \underbrace{ \nabla_{x_t} \log \left[ \frac{p(x_t|z_s)p(x_t|z_c)}{p(x_t)} \left( \frac{\frac{p(x_t|z_c,z_s)}{p(x_t)}}{\frac{p(x_t|z_c)p(x_t|z_s)}{p(x_t)^2}} \right) \right] }_{\text{Eq. (29)}} + \cancelto{0}{\nabla_{x_t} \log p(z_c,z_s)},
\end{align*}

where $\nabla_{x_t} \log p(z_c,z_s)=0$ because it is constant w.r.t. $x_t$.

\clearpage

\section{Derivation for Corollary 1}
\label{sec:derivation for corollary 3.3}


The derivation starts from GCDM formulation proposed in Definition 1 in the main paper.

\begin{align}
    \nabla_{x_t} \log \tilde{p}_{\alpha, \lambda, \beta_c, \beta_s}(x_t|z_c,z_s) \triangleq \,\,
    & \epsilon(x_t,t) + \alpha \Bigl[ \lambda (\underbrace{\epsilon(x_t, t, z_c, z_s) - \epsilon(x_t,t)}_{\nabla_{x_t} \log p(z_c,z_s|x_t)}) \\
    & + (1-\lambda) \sum_{i=\{c,s\}} \beta_{i} \Bigl( \underbrace{\epsilon(x_t,t,z_i)- \epsilon(x_t,t)}_{\nabla_{x_t} \log p(z_i|x_t)} \Bigr) \Bigr] \,. \nonumber
\end{align}

\noindent
Given the fact that $\epsilon(x_t, t)=\nabla_{x_t} \log p(x_t)$, taking integral w.r.t. $x_t$ to the equation yields:

\begin{align}
    \log \tilde{p}_{\alpha, \lambda, \beta_c, \beta_s}(x_t|z_c,z_s) = &
    \log p(x_t) + \alpha \Bigl[ \lambda (\log p (x_t | z_c, z_s) - \log p(x_t)) \\
    & + (1-\lambda) \sum_{i=\{c,s\}} \beta_{i} \Bigl( \log p(x_t | z_i)- \log p (x_t) \Bigr) \Bigr] + C \,, \nonumber
\end{align}

\noindent where $C$ is a constant. Merging all the terms with $\log$:

\begin{align}
    \log \, & \tilde{p}_{\alpha, \lambda, \beta_c, \beta_s}(x_t|z_c,z_s) = \nonumber \\ 
    & \log \text{exp}(C) + \log \Bigl( p(x_t) \left( \frac{p(x_t | z_c, z_s)}{p(x_t)} \right) ^ {\alpha \lambda} \left( \frac{p(x_t | z_c)^{\beta_c}p(x_t | z_s)^{\beta_s}}{p(x_t)^{\beta_c+\beta_s}} \right)^{\alpha (1-\lambda)} \Bigr) \,
\end{align}

\noindent Taking exponential to the above equation:

\begin{align}
    & \tilde{p}_{\alpha, \lambda, \beta_c, \beta_s}(x_t|z_c,z_s) \nonumber \\
    & = \text{exp}(C) p(x_t) \left( \frac{p(x_t | z_c, z_s)}{p(x_t)} \right) ^ {\alpha \lambda} \left( \frac{p(x_t | z_c)^{\beta_c}p(x_t | z_s)^{\beta_s}}{p(x_t)^{\beta_c+\beta_s}} \right)^{\alpha (1-\lambda)} \, \\
    & = \text{exp}(C) p(x_t)^{(1 - \alpha\lambda - \alpha(1-\lambda)(\beta_c + \beta_s))} p(x_t | z_c, z_s) ^ {\alpha \lambda} \left( p(x_t | z_c)^{\beta_c} p(x_t | z_s)^{\beta_s} \right)^{\alpha (1-\lambda)} \,. 
\end{align}

\noindent Given the fact that $\beta_c + \beta_s = 1$,

\begin{align}
    & \tilde{p}_{\alpha, \lambda, \beta_c, \beta_s}(x_t|z_c,z_s) \nonumber \\
    & = \text{exp}(C) p(x_t)^{(1-\alpha)} \left[ p(x_t | z_c, z_s) ^ {\lambda}  \left( p(x_t | z_c)^{\beta_c} p(x_t | z_s)^{(1-\beta_c)} \right)^{(1-\lambda)} \right]^{\alpha} \,.
\end{align}

\noindent Since the exponential function is always positive, 

\begin{align}
    \tilde{p}_{\alpha, \lambda, \beta_c, \beta_s}(x_t|z_c,z_s) \propto p(x_t)^{(1-\alpha)} \left[ p(x_t | z_c, z_s) ^ {\lambda}  \left( p(x_t | z_c)^{\beta_c} p(x_t | z_s)^{(1-\beta_c)} \right)^{(1-\lambda)} \right]^{\alpha} \,,
\end{align}

\noindent which is the same as Corollary 1 in the main paper. 

\section{Preliminaries on Diffusion Models}
\label{sec:preliminaries on diffusion models}
Diffusion Models~\citep{sohl2015deep,ho2020denoising} are one class of generative models that map the complex real distribution to the simple known distribution. In high level, DMs aim to train the networks that learn to denoise a given noised image and a timestep $t$. The noised image is obtained by a fixed noising schedule.
Diffusion Models~\citep{sohl2015deep,ho2020denoising} are formulated as $p_{\theta}(x_0)$. The marginal $p_{\theta}(x_0)$ can be formulated as a marginalization of the joint $p_\theta(x_{0:T})$ over the variables $x_{1:T}$, where $x_1,...x_T$ are latent variables, and $p(x_T)$ is defined as standard gaussian. Variational bound of negative log likelihood of $p_{\theta}(x_0)$ can be computed by introducing the posterior distribution $q(x_{1:T}|x_0)$ with the joint $p_\theta(x_{0:T})$. 
In Diffusion Models~\citep{sohl2015deep,ho2020denoising}, the forward process $q(x_{1:T}|x_0)$ is a predefined Markov Chain involving gradual addition of noise sampled from standard Gaussian to an image. Hence, the forward process can be thought of as a fixed noise scheduler with the $t$-th factorized component $q(x_t|x_{t-1})$ represented as: $q(x_t|x_{t-1})=\mathcal{N}(x_t;\sqrt{1-\beta_t}x_{t-1}, \beta_tI)$, where $\beta_t$ is defined manually. On the other hand, the reverse or the generative process $p_\theta(x_{0:T})$ is modelled as a denoising neural network trained to remove noise gradually at each step. The $t$-th factorized component $p_{\theta}(x_{t-1}|x_t)$ of the reverse process is then defined as, $\mathcal{N}(x_{t-1}|\mu_{\theta}(x_t,t),\Sigma(x_t,t))$. Assuming that variance is fixed, the objective of Diffusion Models (estimating $\mu$ and $\epsilon$) can be derived using the variational bound~\citep{sohl2015deep, ho2020denoising} (Refer the original papers for further details).

Following Denoising Diffusion Probabilistic Models~\citep{ho2020denoising} (DDPM), Denoising Diffusion Implicit Model~\citep{song2020denoising} (DDIM) was proposed that significantly reduced the sampling time by deriving a non-Markovian diffusion process that generalizes DDPM. The latent space of DDPM and DDIM has the same capacity as the original image making it computationally expensive and memory intensive. Latent Diffusion Models~\citep{rombach2022high} (LDM) used a pretrained autoencoder~\citep{esser2021taming} to reduce the dimension of images to a lower capacity space and trained a diffusion model on the latent space of the autoencoder, reducing time and memory complexity significantly without loss in quality.

All our experiments are based on LDM as the base diffusion model with DDIM for sampling. However the techniques are equivalently applicable to any diffusion model and sampling strategy.

\section{Implementation Details}
\label{sec:details}

We build our models on top of LDM codebase\footnote{\url{https://github.com/CompVis/latent-diffusion}}. For FFHQ and LSUN-church, we train our model for two days with eight V--100 GPUs. The model for AFHQ dataset is trained for one and a half days with the same device. All models are trained for approximately 200000 iterations with a batch size of 32, 4 samples per GPU without gradient accumulation. All models are trained with 256$\times$256 images with a latent $z$ size of 3$\times$64$\times$64. The dimensions of content code $z_c$ is 1$\times$8$\times$8 while that of style code $z_s$ is 512$\times$1$\times$1. $t_1,t_2$ and $t_3$ from Eq. 1 in the main paper are timestep embeddings learned to specialize according to the latent code they are applied for to support learning different behavior for content and style features at different timesteps.
We also experimented with different sizes for content and style code and chose these for best empirical performance. The content encoder takes as input $z$ and outputs $z_c$ following a sequence of ResNet blocks. The style encoder has a similar sequence of ResNet blocks followed by a final global average pooling layer to squish the spatial dimensions similar to the semantic encoder in~\citep{preechakul2022diffusion}. 

To support GCDM during sampling, we require the model to be able to generate meaningful scores and model the style, content and joint distributions. Hence, during training we provide only style code, only content code and both style and content code all with probability 0.3 (adding up to 0.9) and no conditioning with probability 0.1 following classifier--free guidance literature. This helps learn the conditional and unconditional models that are required to use the proposed GCDM formulation. The code will be released upon acceptance of the paper.

During sampling, without \emph{reverse DDIM}, if all the joint, conditionals, and unconditional guidance are used, sampling time for a single image is 10 seconds. With \emph{reverse DDIM} to get $x_T$ where T is the final timestep, it takes 22 seconds. This might be lesser if \emph{reverse DDIM} is stopped early and generation happens from the stopped point. Specific hyperparameters used to generate results in the main paper and appendix are provided in Table.~\ref{table:used hyperparameters for the figures}.
\begin{table}[h]
        \vspace{-0.5em}
        \captionof{table}{Hyperparameters used to generate the figures in the main paper and appendix. Timestep scheduling is only used in the sampling process. The parenthesis in the second column indicates the number of steps we used for sampling. Note that $\beta_c=1-\beta_s$.}
\centering
    \begin{adjustbox}{width=0.9\textwidth}
    \begin{tabular}{c c c c c c c c c}
    \toprule
     &  \multicolumn{7}{c}{Main paper}\\
    \midrule
    \textbf{Dataset} & \textbf{sampler} & \bm{$x_T$} & \bm{$\alpha$} & \bm{$\lambda$} & \bm{$\beta_s$} & \bm{$a$} & \bm{$b$} & \textbf{scheduler} \\  
    FFHQ & DDIM+SDEdit (60) & \emph{reverse DDIM} & 1.5 & 0.9 & 1.0 & 0.025 & 550 & sigmoid \\
    LSUN-church & DDIM (100) & $q(x_{991}|x_0)$ & 2.0 & 0.5 & 0.0 & - & - & - \\
    AFHQ & DDIM+SDEdit (60) & $q(x_{591}|x_0)$ & 3.0 & 0.75 & 1.0 & - & - & - \\ 
    
    \midrule
    &  \multicolumn{7}{c}{Appendix}\\

    \midrule
    FFHQ & DDIM (100) & \emph{reverse DDIM} & 1.5 & 0.9 & 1.0 & 0.025 & 550 & sigmoid \\
    LSUN-church & DDIM (100) & \emph{reverse DDIM} & 5.0 & 0.5 & 0.0 & 0.025 & 600 & sigmoid \\
   \bottomrule
   
    \end{tabular}
    \end{adjustbox}
    \label{table:used hyperparameters for the figures}
\end{table}

\begin{figure*}[h]
\includegraphics[width=\textwidth]{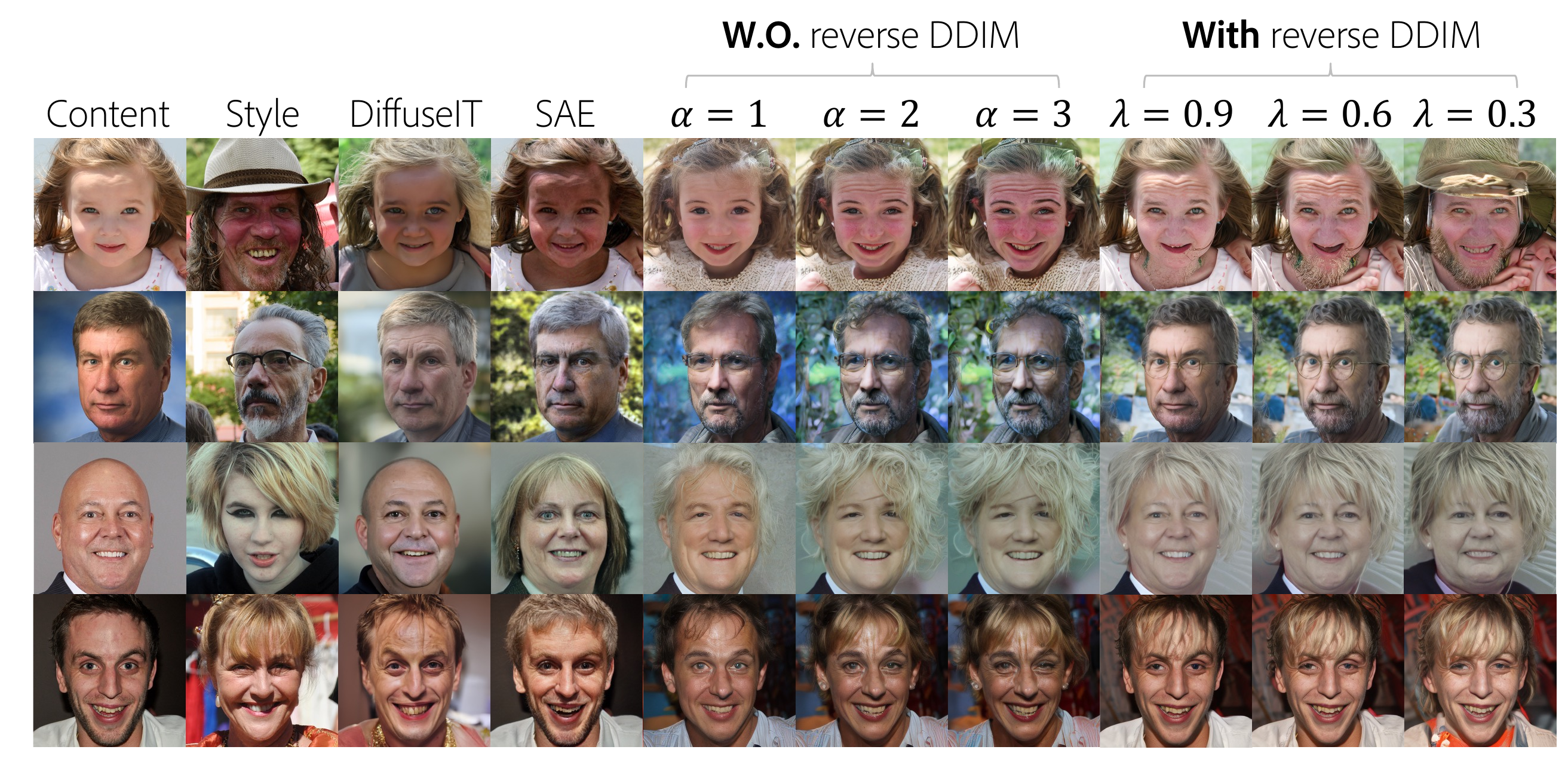}
\caption{Comparisons between with and without \emph{reverse DDIM} sampling for FFHQ model. We use the proposed GCDM and timestep scheduling during sampling. We can clearly see better identity preservation than not using reverse DDIM, on par with SAE~\citep{park2020swapping} while still providing better controllability.}
\label{fig:FFHQ fig4}
\end{figure*}

\subsection{Experimental Setup}
\label{sub:exp setup}

\noindent\textbf{Datasets}\\
We train different models on the commonly used datasets such as AFHQ~\citep{choi2020stargan}, FFHQ~\citep{karras2019style} and LSUN-church~\citep{yu2015lsun}. 



\noindent\textbf{Baselines} \\
\indent\textbf{DiffuseIT:} The most similar work to ours based on diffusion models is DiffuseIT~\citep{kwon2022diffusion} that tackles the same problem formulation. We compare our results with DiffuseIT using their pretrained model and default parameters. \\
\indent\textbf{DiffAE+SDEdit:} Since Diffusion Autoencoder~\citep{preechakul2022diffusion} does not directly support image-to-image translation, we combine that with SDEdit~\citep{meng2021sdedit}. The input image for the reverse process is $x_{600}$ (chosen empirically) obtained as $q(x_{600}|x_c)$ by running the forward process on the content image. The semantic feature $z_{sem}$ from the semantic encoder of DiffAE is used given the style image $x_s$.\\ 
\indent\textbf{DiffAE+MagicMix:} We also combine MagixMix~\citep{liew2022magicmix} with DiffAE. Similar to DiffAE+SDEdit, this model takes $x_{600}$ from $x_c$ as input and $z_{sem}$ from $x_s$ as conditioning. Additionally, at each timestep, the approximated previous timestep $\hat{x}_{t-1}$ is combined with $x_{t-1}$ from the content image $x_c$, i.e., $\hat{x}_{t-1}=v\hat{x}_{t-1} + (1-v)q(x_{t-1}|x_c)$. For this experiment, $v=0.5$ is used and the noise mixing technique is applied between $t=[600,300]$. \\
\indent\textbf{SAE:} Swapping Autoencoder~\citep{park2020swapping} based on GAN~\citep{NIPS2014_5ca3e9b1} is also evaluated. Since the available pretrained model is on a resolution of 512, we resize the generated results to 256 for a fair comparison. \\
\indent\textbf{DEADiff:}
The sampling method of DEADiff~\citep{qi2024deadiff} only uses the style representation (Fig. 2 in their paper). To see the performance of both style and content, we give both representations during the inference stage (Note that Fig. 7 in their paper is not from the content representation but from depth ControlNet~\citep{zhang2023adding}). We did not finetune the model specifically for facial data since adjusting the facial dataset to their training data settings is infeasible (e.g., two images with the same style but with a different subject, and two images with the same subject but with a different style). \\
\indent\textbf{StarGAN v2:} The training and sampling process of StarGAN v2~\citep{choi2020stargan} requires gender attributes of the image. Since FFHQ does not have attribute labels, we use their pretrained models on CelebA-HQ. When sampling, we preprocess and label the test images by using a pretrained gender classifier (c.f. EasyFace).

\noindent The training and sampling process of StarGAN v2 requires gender attributes of the image. Since FFHQ does not have attribute labels, we use their pretrained models on CelebA-HQ. When sampling, we preprocess and label the test images by using a pretrained gender classifier (c.f. EasyFace).

\noindent\textbf{Evaluation Metrics} \\
\indent\textbf{FID:} We use the commonly used Fréchet inception distance (FID)~\citep{heusel2017gans} to ensure the generated samples are realistic. We follow the protocol proposed in~\citep{choi2020stargan} for reference-based image translation. To obtain statistics from generated images, 2000 test samples are used as the content images, and five randomly chosen images from the rest of the test set are used as style images for each content image to generate 10,000 synthetic images. \\
\indent\textbf{LPIPS:} Even though FID evaluates the realism of the generations, the model could use just content and ignore style (or vice versa) and still get good FID. Following~\citep{choi2020stargan}, we use LPIPS score obtained by measuring the feature distances between pairs of synthetic images generated from the same content image but with different style images. \textbf{Higher LPIPS indicates more diverse results}. Ideally, the models incorporate enough style information from different style images for the same content image (increasing LPIPS) but without going out of the real distribution (decreasing FID). \\

\subsection{Identity Preservation}
\label{subsec:reverse_ddim}
We notice that when the proposed sampling technique is used with randomly sampled noise $x_T$ for reference based image translation or manipulation, particularly on FFHQ dataset, that the identity of the content image is not preserved. This is an important aspect of image manipulation for faces. One of the ways to preserve better identity is to use the deteministic \emph{reverse DDIM} process described in~\citep{preechakul2022diffusion} to obtain $x_T$ that corresponds to a given content image. To do this, we pass the content image to both the content and style encoders as well as the diffusion model to get $x_T$ that reconstructs the content image. This $x_T$ is then used along with content code from content image and style code from style image to generate identity preserving translation.

The comparisons between with and without \emph{reverse DDIM} during sampling are provided in Table~\ref{table:FFHQ comparisons between with and without reverse ddim}. We can see that with \emph{reverse DDIM}, we can expect to get more realistic results (lower FID) while the less diverse results (lower LPIPS). Fig.~\ref{fig:FFHQ fig4} also shows comparisons between with and without \emph{reverse DDIM}. In contrast to SAE~\citep{park2020swapping} that preserves better identity by trading of magnitude of style applied, our approach provides the ability to control the magnitude of identity preservation and style transfer independently.

\begin{figure}[h]
\centering
\begin{minipage}{.45\linewidth}
  \includegraphics[width=\textwidth]{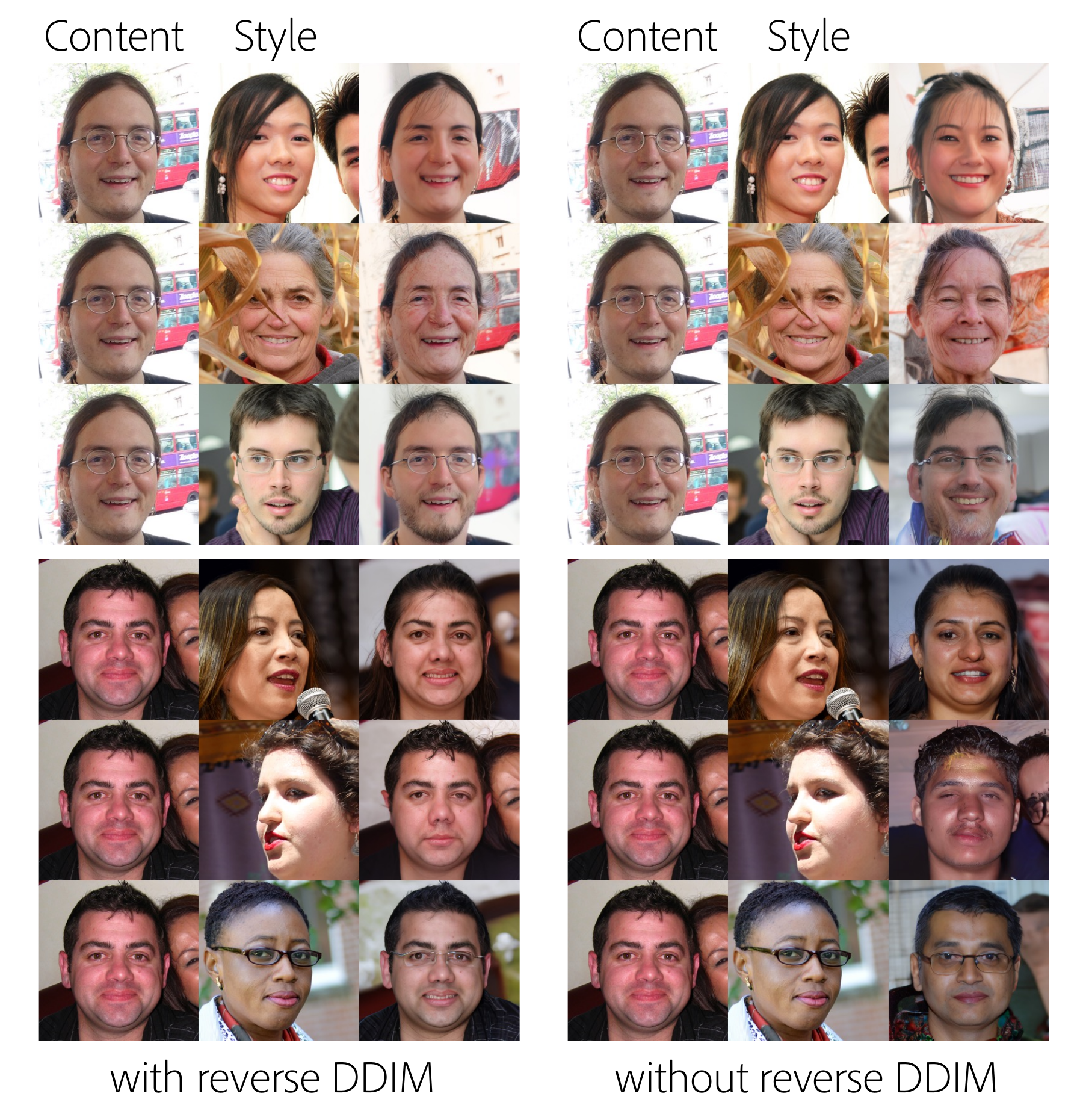}
  \captionof{figure}{Comparisons between with and without DDIM reverse sampling method in FFHQ dataset.}
  \label{fig:FFHQ multi c multi s}
\end{minipage}
\hspace{.05\linewidth}
\begin{minipage}{.45\linewidth}
  \includegraphics[width=\textwidth]{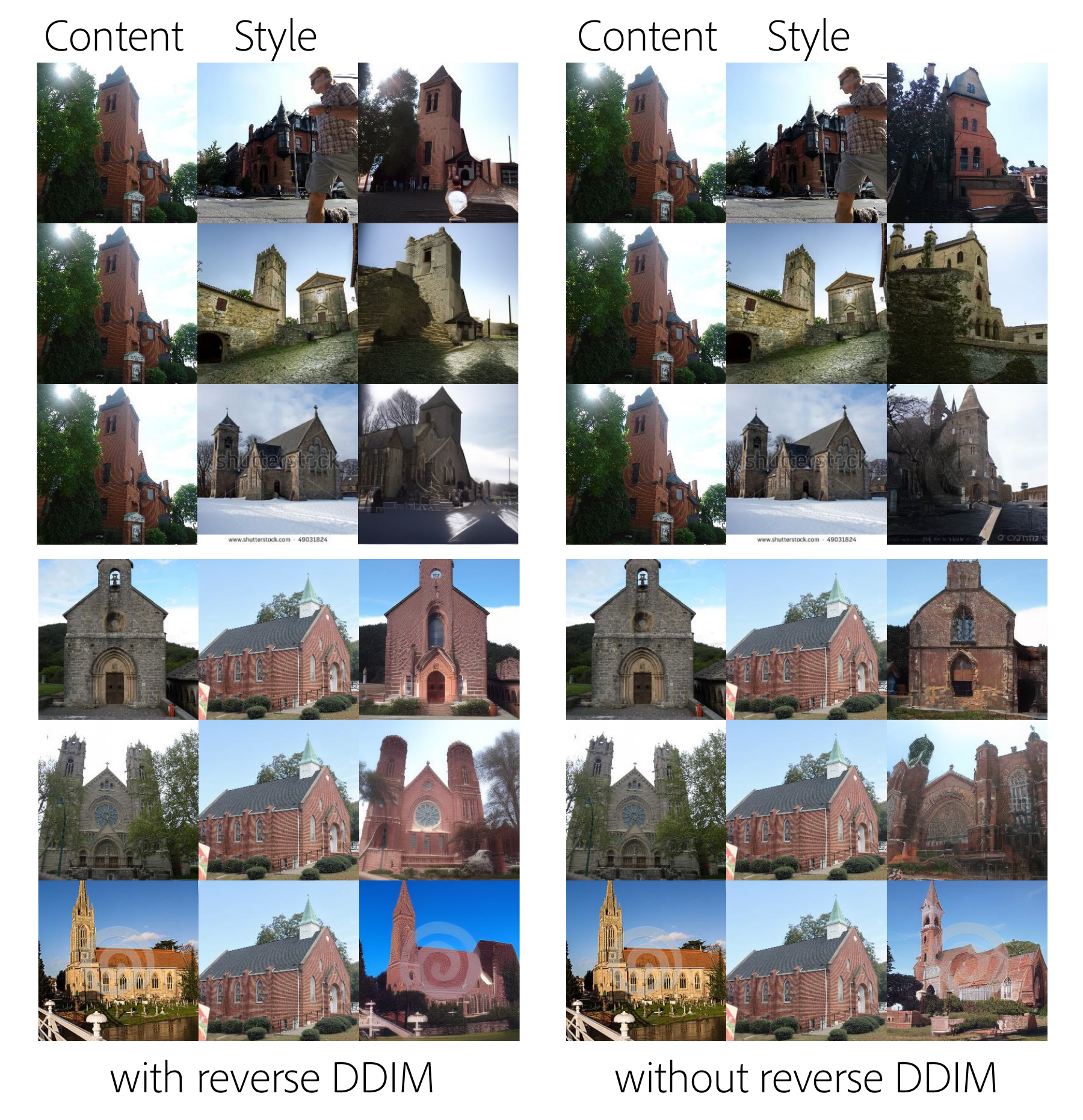}
  \captionof{figure}{Comparisons between with and without DDIM reverse sampling method in LSUN-church dataset.}
  \label{fig:LSUN churches multi c multi s}
\end{minipage}
\end{figure}

Additional comparisons between with and without \emph{reverse DDIM} are provided in Fig.~\ref{fig:FFHQ multi c multi s} and \ref{fig:LSUN churches multi c multi s}. We can see that the results with \emph{reverse DDIM} better preserve the content identity while applying the style reasonably. On the other hand, the results without \emph{reverse DDIM} have a stronger impact on style with lesser identity preservation, which could be preferable in non-face domains such as abstract or artistic images or for semantic mixing. 

\begin{table*}[!t]
        \vspace{-0.5em}
        \captionof{table}{Quantitative comparison between the variants of ours with and without \emph{reverse ddim}.}
\centering
    \begin{adjustbox}{width=0.9\textwidth}
    \begin{tabular}{c c c c c c c}
    \hline 
    & \multicolumn{3}{c}{w/o \emph{reverse ddim}} & \multicolumn{3}{c}{w/ \emph{reverse ddim}} \\
    \cmidrule(lr){2-4}
    \cmidrule(lr){5-7}
    & Ours($\alpha=1.0$) & Ours($\alpha=2.0$) & Ours($\alpha=3.0$) & Ours($\lambda=0.9$) & Ours($\lambda=0.6$) & Ours($\lambda=0.3$) \\ 
   FID & 20.38 & 23.68 & 26.45 & \textbf{11.99} & 13.40 & 15.45  \\ 
   LPIPS & 0.53 & 0.57 & \textbf{0.6} & 0.34 & 0.42 & 0.49 \\
   \hline
    \end{tabular}
    \end{adjustbox}
    \vspace{-0.5em}
    \label{table:FFHQ comparisons between with and without reverse ddim}
\end{table*}

\clearpage
\section{Extent of Controllability}
\label{sec:controls}

In this Section, we present the rich controllability of our proposed framework. The latent space exploration by PCA is presented in the main paper. Interpolation results are shown in Section~\ref{subsec:interpolation}. Further analysis of the latent space by KNN is reported in Section~\ref{subsec:knn}.

\subsection{Interpolation}
\label{subsec:interpolation}

We conducted experiments on latent space interpolation in order to analyze the effects of the content, the style, and $x_T$ during the sampling process. All the results use \emph{reverse DDIM} with content image to get $x_T$.

\begin{figure*}[!h]
\includegraphics[width=\textwidth]{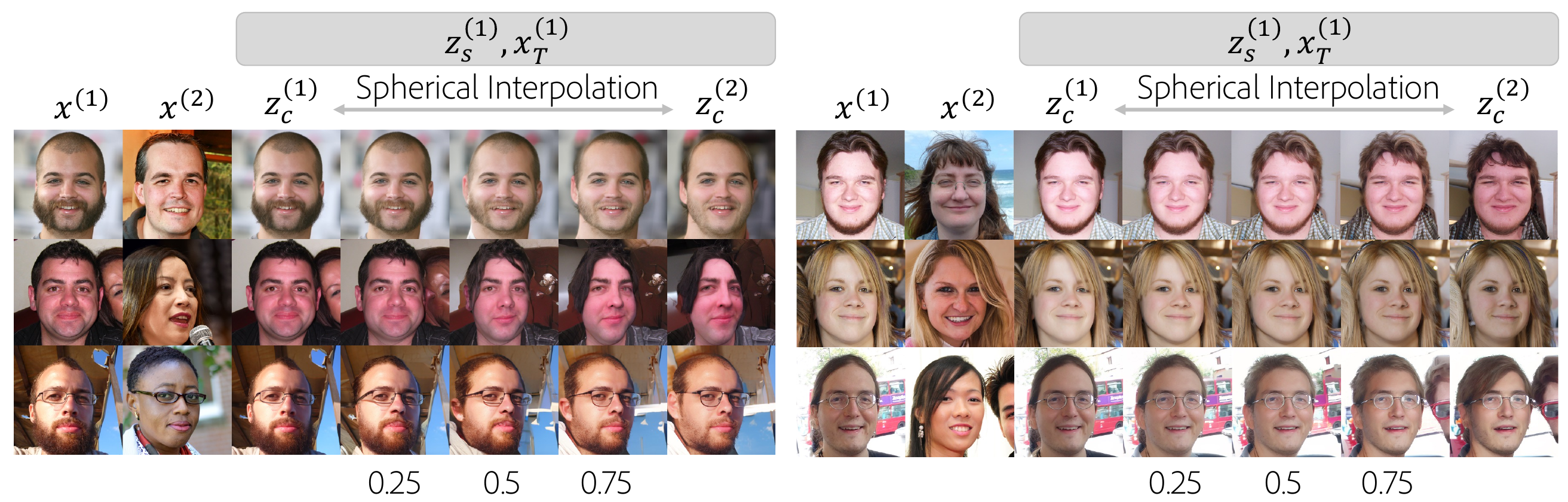}
\caption{Content interpolation results. Style and $x_T$ are obtained from images in the first column while content code is interpolated between images in column 1 and column 2. We can see how content-specific factors vary smoothly.}
\label{fig:FFHQ content interp}
\end{figure*}

Fig.~\ref{fig:FFHQ content interp} shows the content-only interpolation results where style feature $z_s$ and noise $x_T$ are fixed to the image in the first column. The gray box on the top indicates the fixed input while $z_c$ is interpolated between the two images in the first two columns. From the figure, we can see that the style information and the person's identity are maintained while pose and facial shape are changed.

\begin{figure*}[!h]
\includegraphics[width=\textwidth]{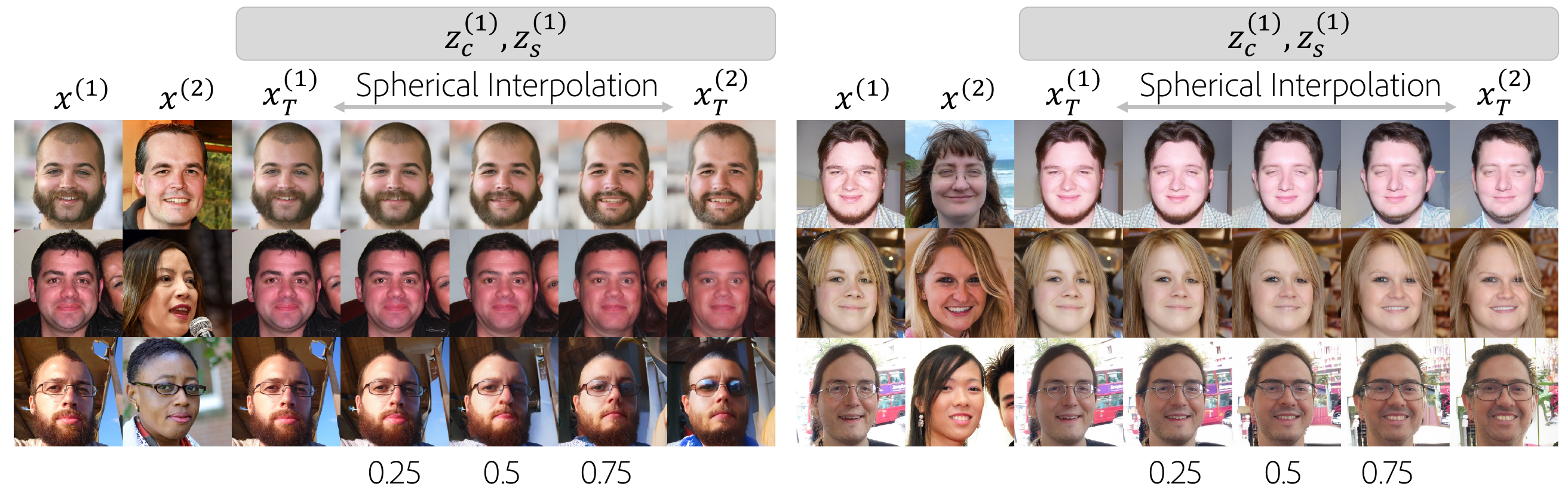}
\caption{$x_T$ interpolation results where style and content are obtained from images in the first column while $x_T$ is interpolated between reverse DDIM of both images. We can see stochastic changes causing mild identity changes. Fixing $x_T$ to the content image hence provides better identity preservation for image translation and manipulation.}
\label{fig:FFHQ xT interp}
\end{figure*}

Fig.~\ref{fig:FFHQ xT interp} shows the case $x_T$ obtained from \emph{reverse DDIM} of the images in the first two columns is interpolated while the style and content features are fixed to the image in the first column. The content (e.g., pose, facial shape) and the style (e.g., beard, eyeglasses, and facial color) are maintained while stochastic properties change. We can see that identity is not entirely tied to $x_T$ but the stochasticity causes changes in the identity. This could be an evidence of why using \emph{reverse DDIM} to fix $x_T$ preserves better identity.

\begin{figure*}[!h]
\includegraphics[width=\textwidth]{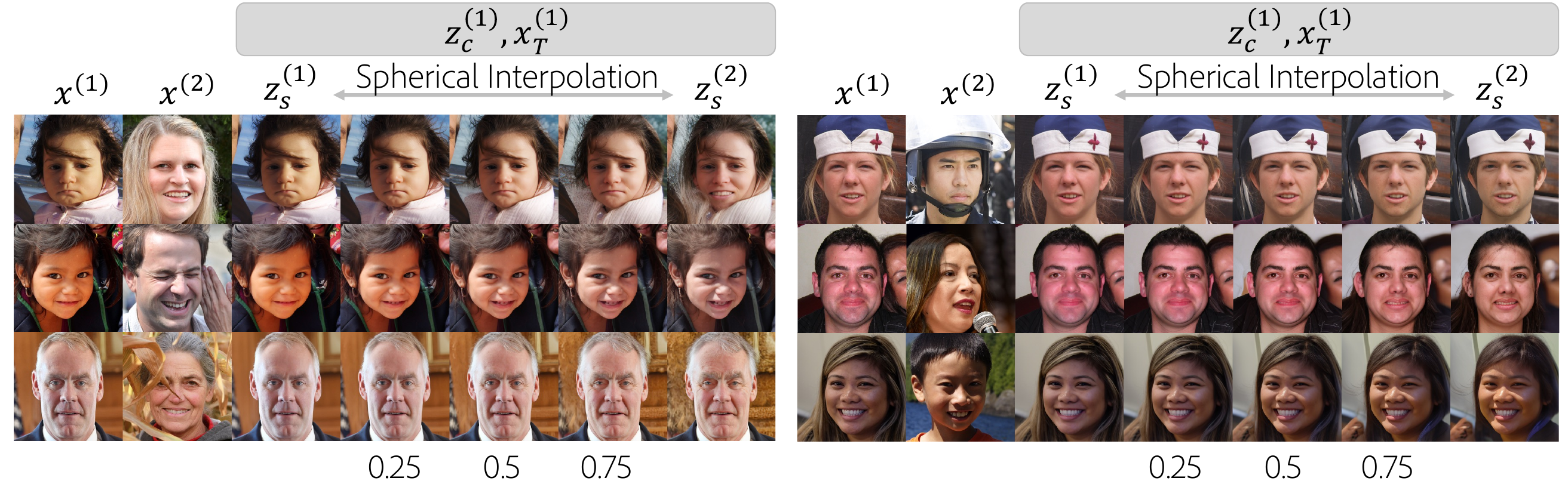}
\caption{Style interpolation results when content and $x_T$ are obtained from images in the first column. We can see smooth changes in the semantic attributes such as age, gender, smile, eyeglasses, etc. allowing for effective style manipulations.}
\label{fig:FFHQ style interp}
\end{figure*}

Fig.~\ref{fig:FFHQ style interp} visualizes the style interpolation while the content and $x_t$ are fixed to the first image. The person's identity, pose, and facial shape are preserved while the facial expression, gender, and age are smoothly changed validating our results.


\subsection{Information Encoded in Each Latent Space}\label{subsec:Information Encoded in Each Latent Space}
\begin{figure}[h]
\includegraphics[width=\textwidth]{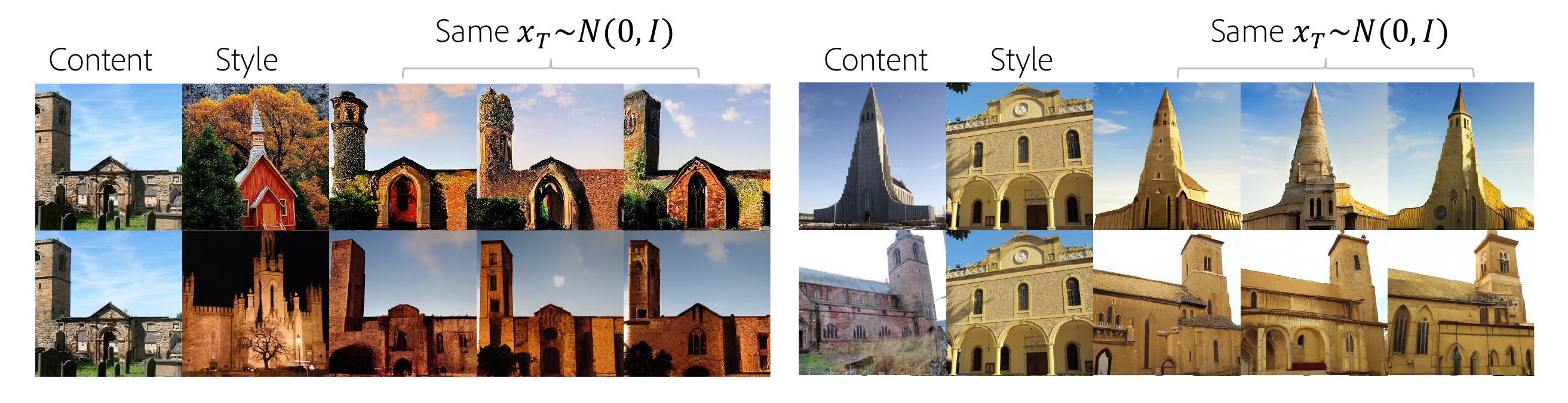}
\caption{Reference-based image translation results on LSUN-church.}
\label{fig:LSUN_churches multiple style with single content}
\end{figure}

\begin{figure}[h]
\centering
\includegraphics[width=0.3\textwidth]{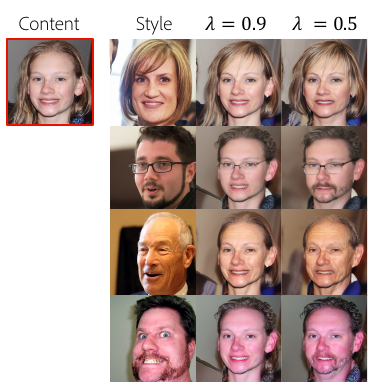}
\caption{Visualizations of a wide spectrum of our generation given a fixed content with different styles.}
\label{fig:various_controls_for_single_content}
\end{figure}

We analyze the role of the denoising network $\epsilon$ and the encoders $E_c$ and $E_s$ by analyzing what information is encoded in the latent spaces. Fig.~\ref{fig:LSUN_churches multiple style with single content} and Fig.~\ref{fig:various_controls_for_single_content} show the results of fixing the content while varying the style images (and vice versa). $x_T$ is fixed as well to reduce the stochasticity. The remaining stochasticity comes from the white noise at each timestep during the reverse process. From the results, we can see that the structure information is maintained while style information changes according to the style image (and vice versa) as we intended. 

\subsection{Interpreting the Latent Spaces based on KNN}
\label{subsec:knn}

\begin{figure*}[b]
\vspace{-0.2in}
\includegraphics[width=\textwidth]{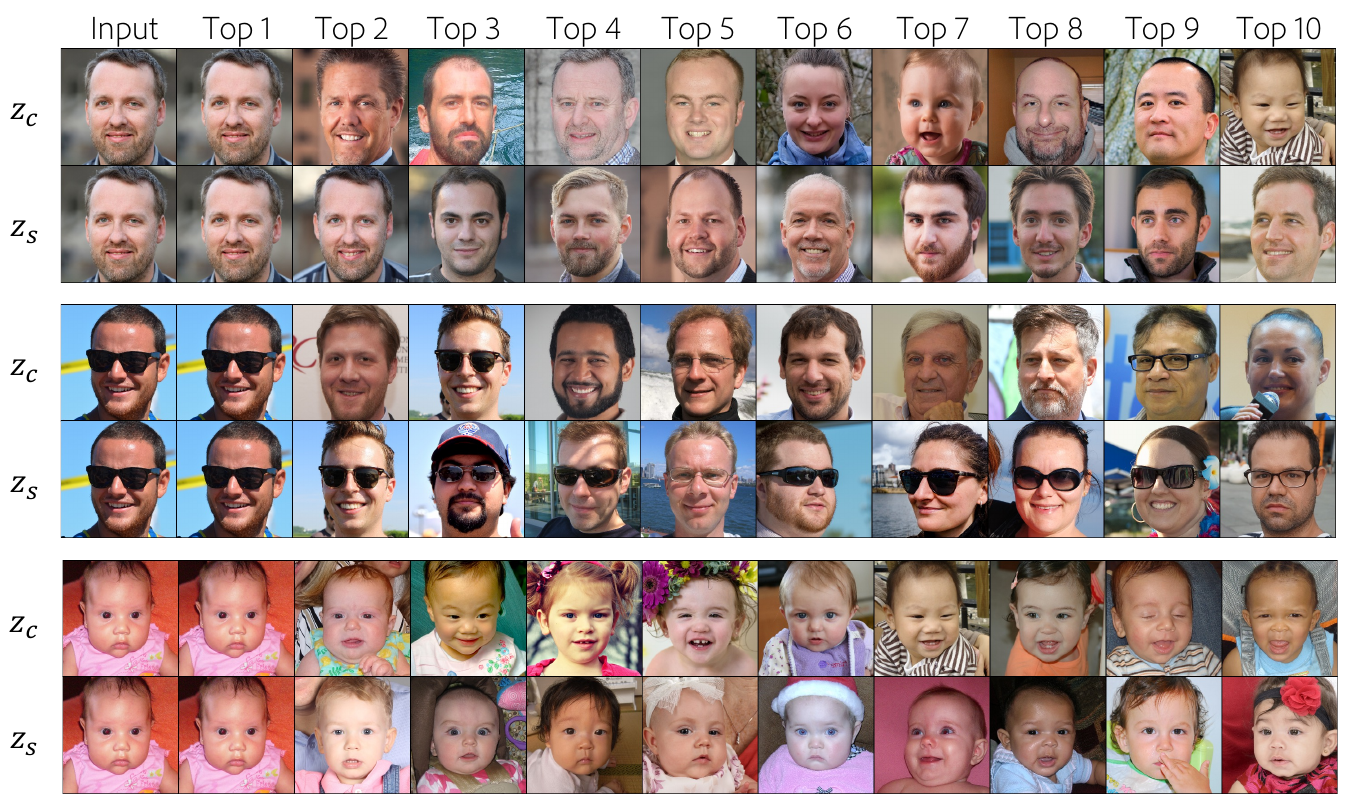}
\vspace{-0.25in}
\caption{KNN results of the content and the style features showing what semantic attributes content and style codes encode.}
\label{fig:FFHQ KNN 1}
\end{figure*}

We additionally perform K Nearest Neighbor (KNN) experiments to understand what features are encoded in the content and style latent representations. We pass 10000 unseen images through the style and the content encoders to get $z_c$ and $z_s$. We then compute the distance of an arbitrary sample with the entire validation set and sort the 10000 distances. 

\begin{figure*}[!h]
\includegraphics[width=\textwidth]{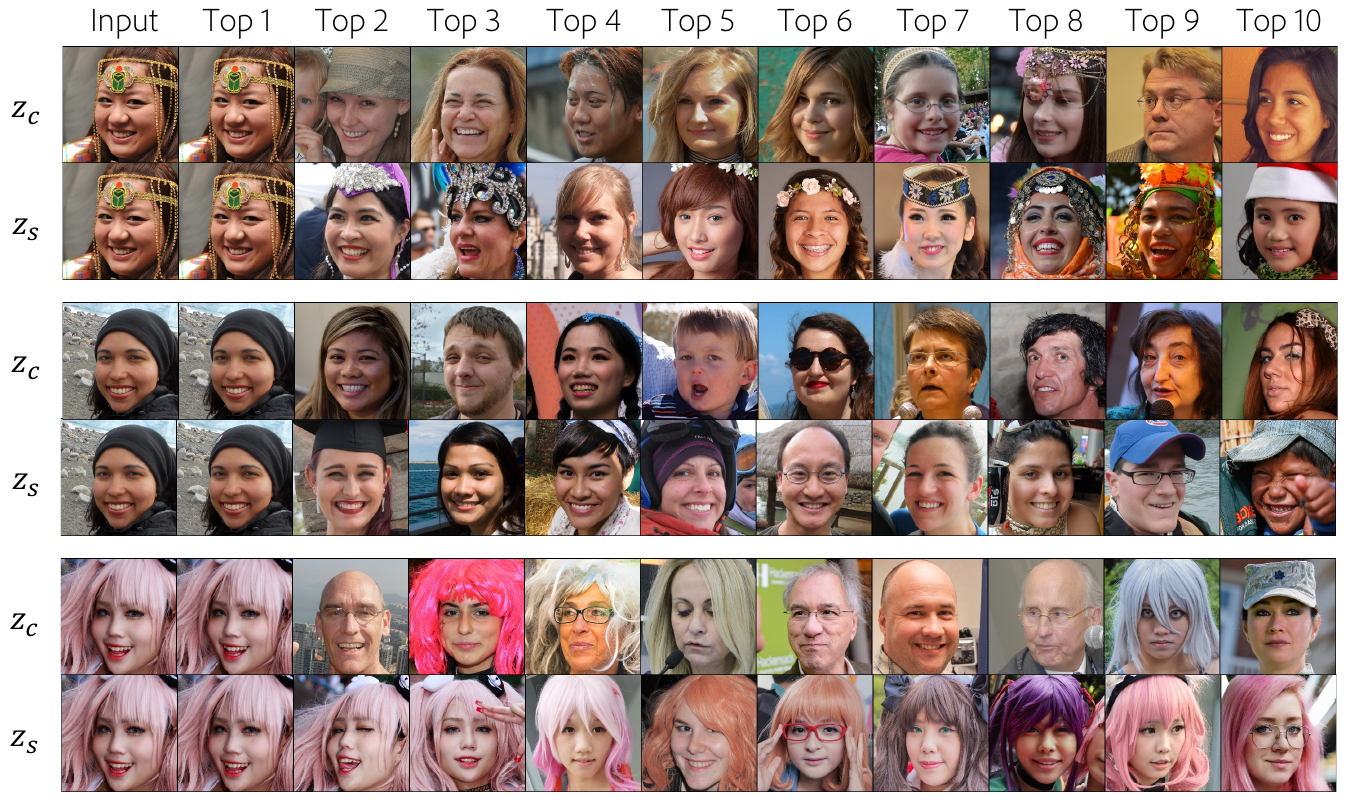}
\caption{KNN results of the content and the style features showing what semantic attributes content and style codes encode.}
\label{fig:FFHQ KNN 2}
\end{figure*}

The results are shown in Fig.~\ref{fig:FFHQ KNN 1} and Fig.~\ref{fig:FFHQ KNN 2}. The first column denotes the input image while the rest of the columns show the top 10 images that have the closest content or style features indicated by $z_c$ (first row within each macro row) and $z_s$ (second row within each macro row) respectively. The second column is the same image. We can see that the content feature mainly contains the pose and the facial shape information while the style has high-level semantics, such as wearing eyeglasses, gender, age, accessories, and hair color.

\clearpage

\section{Timestep Scheduling}
\label{sec:timestep}
\subsection{Training an Implicit Mixture-Of-Experts}
\label{subsec:training timestep}

\begin{figure*}[!h]
\centering
\includegraphics[width=0.5\textwidth]{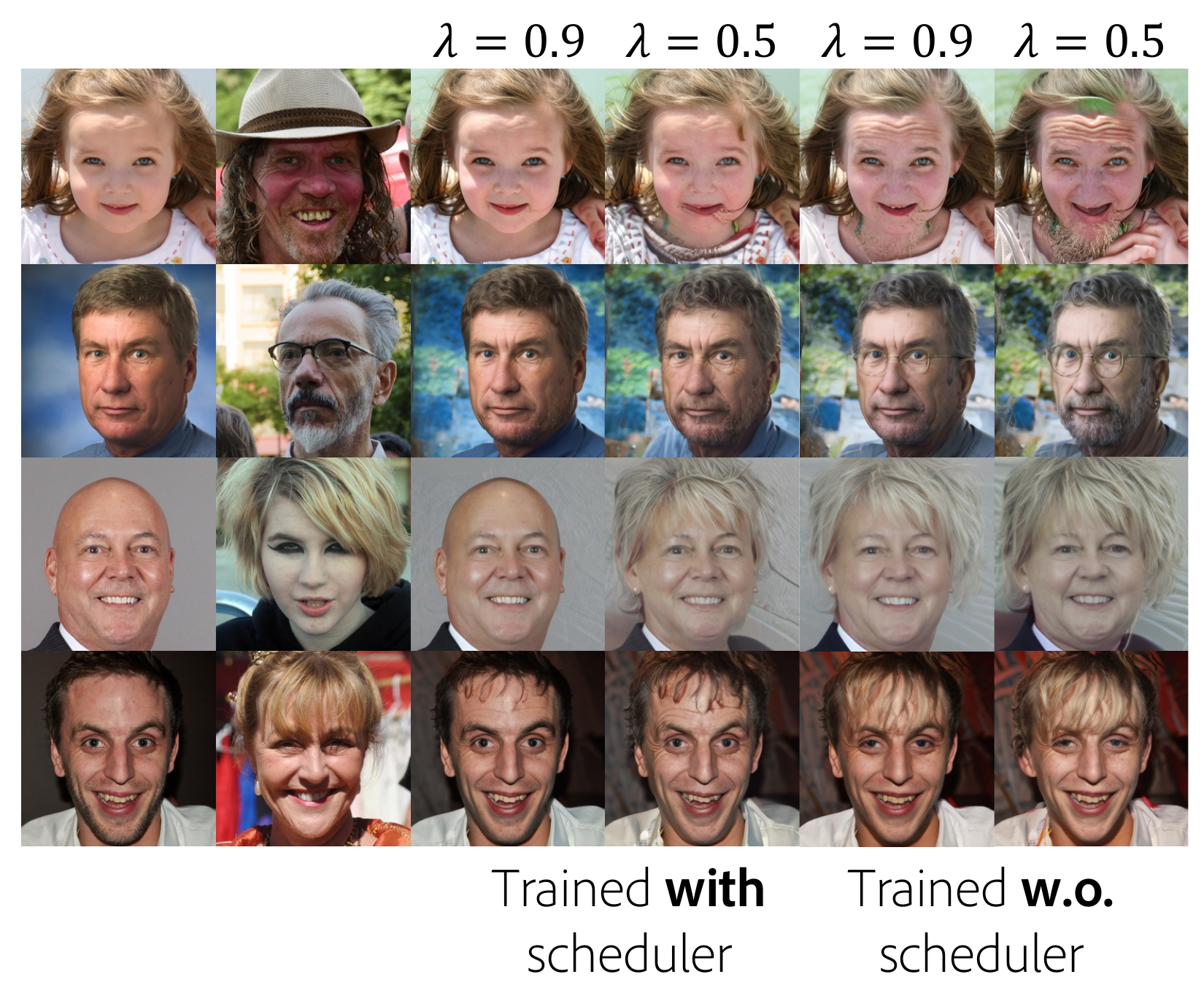}
\caption{Effects of using the proposed timestep scheduling in training.}
\label{fig:FFHQ_trained_with_without_scheduler}
\end{figure*}

Our timestep scheduling approach proposed in Sec 3.2 in the main paper was applied only during sampling for results in the main paper. We trained a model with timestep scheduling applied during training to analyze how it affects the behavior of our framework. Fig.~\ref{fig:timestep scheduling plots} shows the comparisons between the models trained with and without the scheduler. For the results trained with scheduler, we used $a=0.1$ and $b=529$ ($\text{SNR}^{-1}(0.1)$) for both training and sampling. As can be seen in the third and fourth columns (i.e., trained with scheduler), the style effects are relatively small although given $\lambda$ is controlled. It is because the style encoder is trained to be injected only in the early timesteps (0-528), which makes the style representations learn limited features (e.g., eyeglasses are not encoded in the style, as shown in the second row). However, we observe better disentanglement of the content and the style spaces compared to using the timestep scheduling only during sampling. We believe this is because, using timestep scheduling to vary the conditioning input at each timestep implicitly trains the model to specialize to the varied conditioning, implicitly learning a mixture--of--experts like model~\citep{balaji2022ediffi}. We believe this could be a promising avenue for future research.
\subsection{Experiment with different timestep schedules}
\label{subsec:various schedules}
\begin{figure*}[!h]
\includegraphics[width=\textwidth]{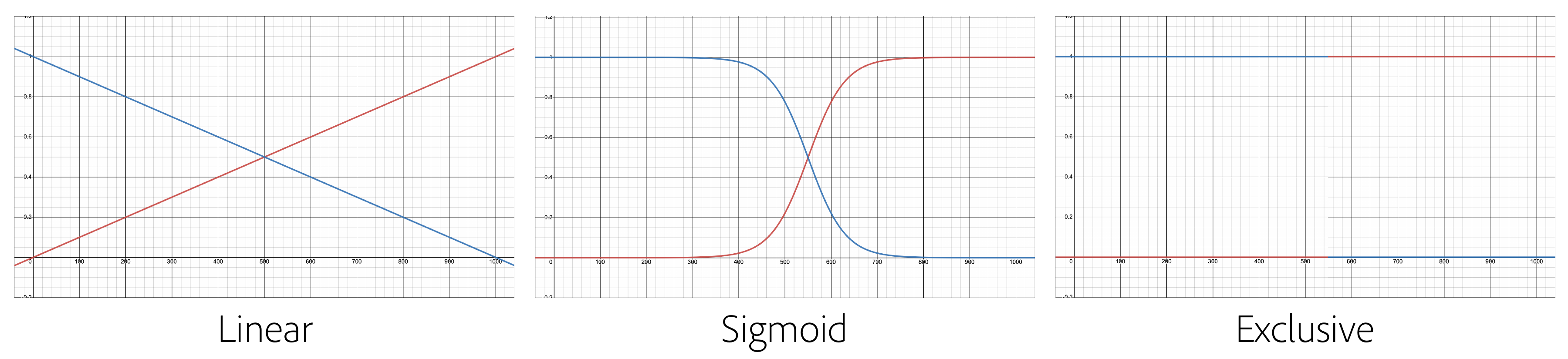}
\caption{Plots for different timestep scheduling strategies. The illustrated plot of the sigmoid scheduler is from $a=0.025$ and $b=550$. Bigger $a$ makes it similar to the exclusive scheduler while smaller $a$ makes it close to the linear scheduler. The blue line indicates the weight scheduler for the style and the red line is for the content.}
\label{fig:timestep scheduling plots}
\end{figure*}
\begin{figure}[h]
\includegraphics[width=\textwidth]{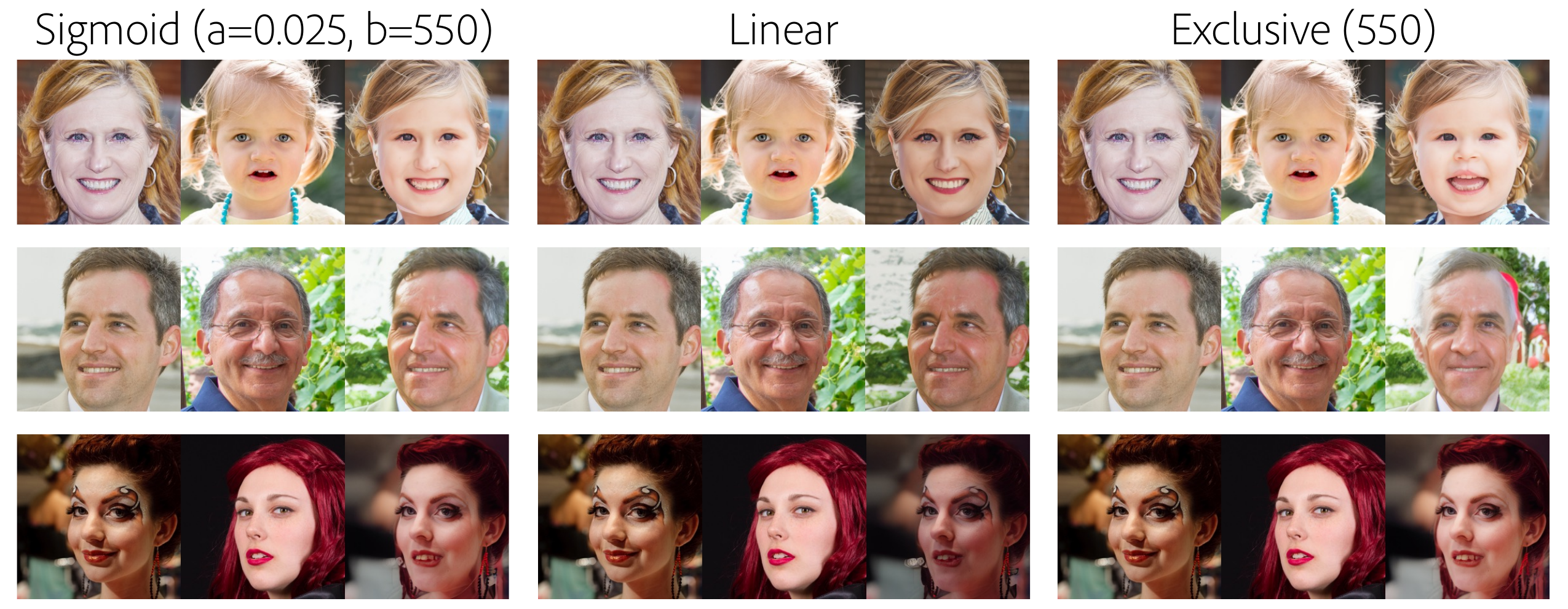}
\caption{Comparisons between different timestep schedulers during sampling. Sigmoid has a softer schedule with more controllability and thus results are more natural generations compared to the other techniques.}
\label{fig:FFHQ timestep scheduling comparisons}
\vspace{-0.2in}
\end{figure}
We compare the different timestep schedulers illustrated in Fig.~\ref{fig:timestep scheduling plots} during sampling. Note that these schedules are not used for training. In the exclusive scheduling, the style weight is one if $t\leq 550$ and zero otherwise. The content weight is applied when style weight is not applied. In the linear scheduling, the style weight linearly decreases from $1$ at $t=0$ to $0$ at $t=999$ while the content weight increases linearly from $0$ to $1$. The sigmoid scheduling is the one proposed in Eq. 11 in the main paper. 

The comparison results are shown in Fig.~\ref{fig:FFHQ timestep scheduling comparisons}. We can observe that the exclusive scheduling shows either magnified style or unnatural generations compared to the sigmoid scheduling. Since it is difficult to exactly define the role of each timestep, naively separating the point where to exclusively apply the content and style yields undesirable results. The linear schedule does not work for all images and has limited control. However, the sigmoid scheduling provides a softer weighting scheme leading to better generations, and has additional controls to get desired results. 

\clearpage

\section{Additional Results}
\label{sec:additional}
\begin{figure}[h]
\centering
\includegraphics[width=0.48\textwidth]{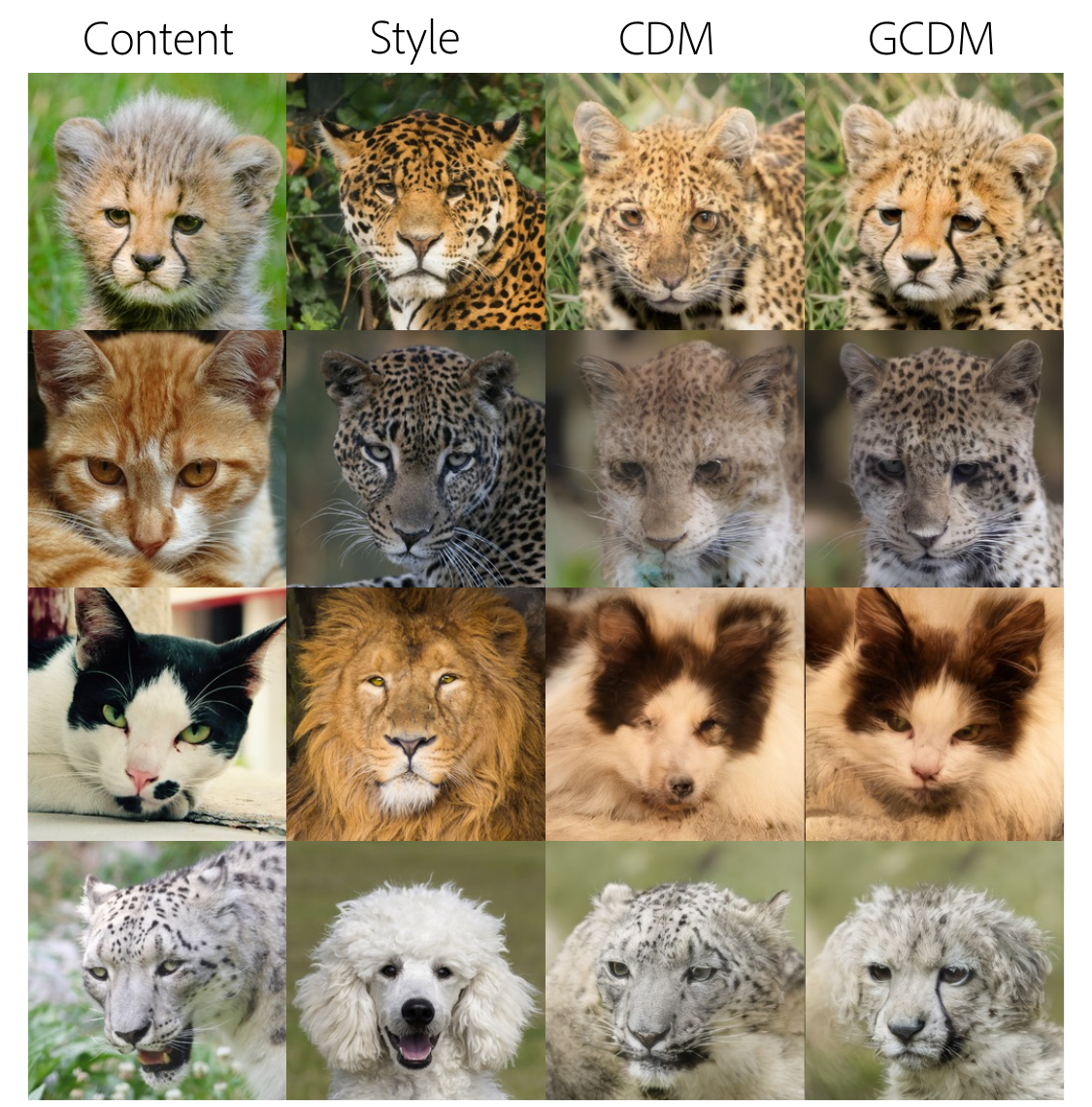}
\caption{\wc{Comparisons between GCDM and CDM demonstrating that CDM can output unnatural images while GCDM can generate realistic images. We use DDIM~\citep{song2020denoising} sampler, and the reverse process is done from $T=600$ inspired by SDEdit~\citep{meng2021sdedit}. $z_{600}$ is obtained by $q(z_{600}|E_{LDM}(x_c))$ using the content image. $x_T$ is randomly sampled.
}}
\label{fig:AFHQ cdm gcdm comparisons}
\end{figure}

In this Section, we provide additional results of our proposed framework. Fig.~\ref{fig:AFHQ cdm gcdm comparisons} shows example generations using CDM and GCDM from the same model. CDM consistently shows worse results than GCDM in reference-based image translation. We can see that by assuming conditional independence, CDM unnaturally overlaps the content and style features.



\begin{figure}[h]
\centering
\includegraphics[width=0.48\textwidth]{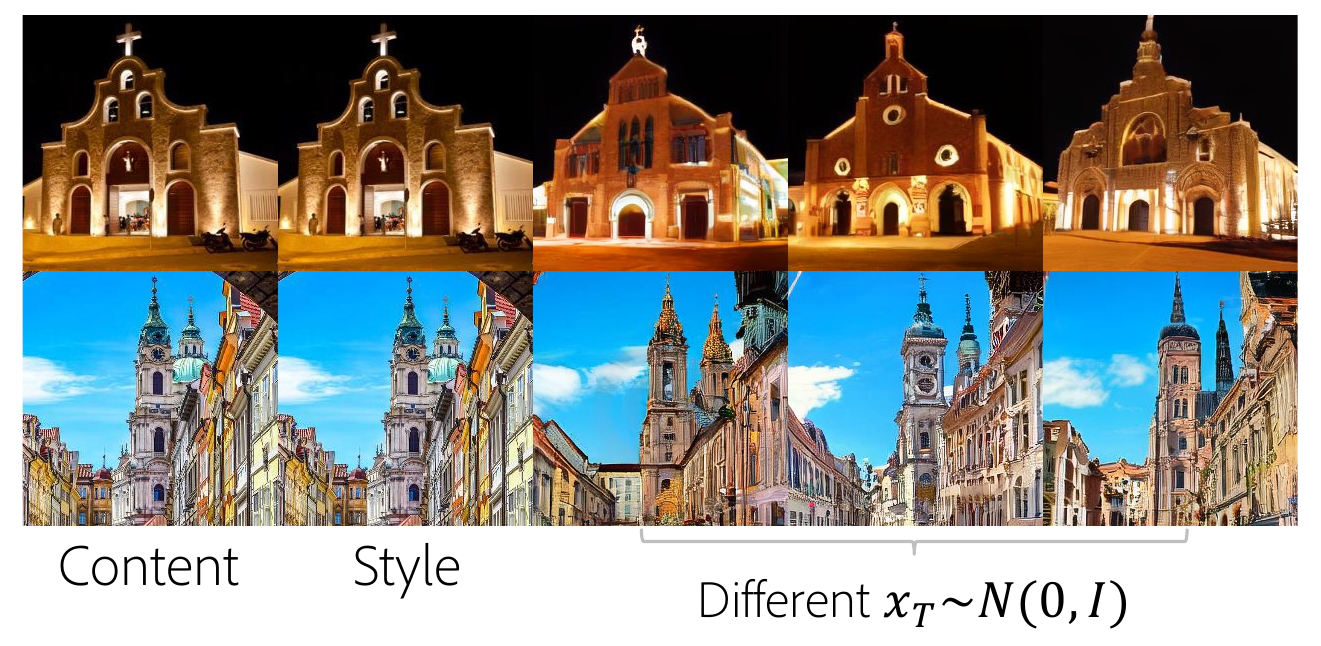}
\caption{Example showing the role of the denoising network during sampling when content and style codes are unchanged. $x_T$ is randomly sampled. The images show that the denoising network play a role in stochasticity since the outputs have consistent shape, color and texture information while minor details of the buildings or clouds are changed.}
\label{fig:LSUN_churches autoencoder}
\vspace{-0.1in}
\end{figure}

\begin{figure*}[h]
\includegraphics[width=\textwidth]{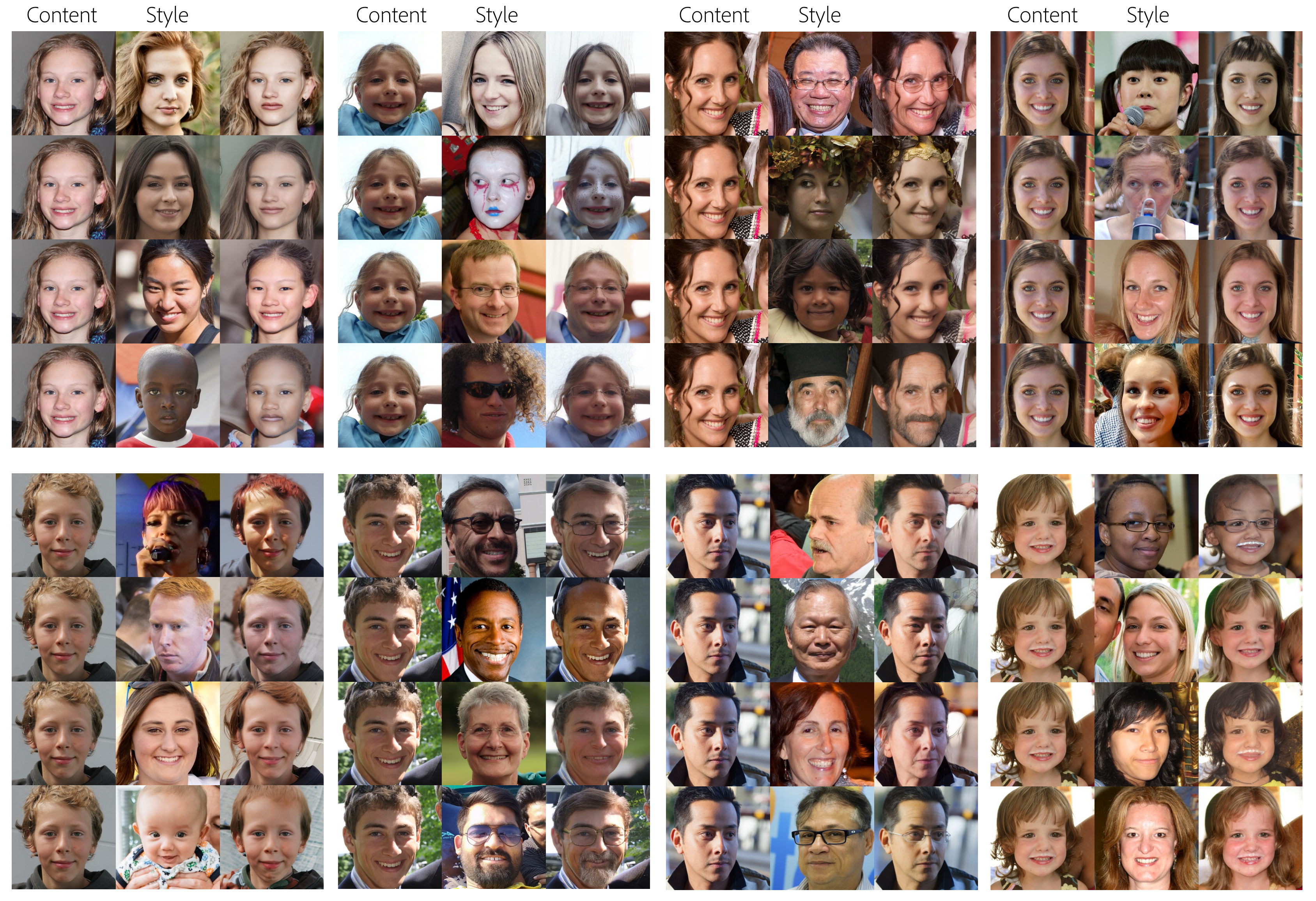}
\caption{Additional results on FFHQ. The results are sampled by \emph{reverse DDIM}.}
\label{fig:FFHQ aditonal results}
\end{figure*}

\begin{figure*}[h]
\includegraphics[width=\textwidth]{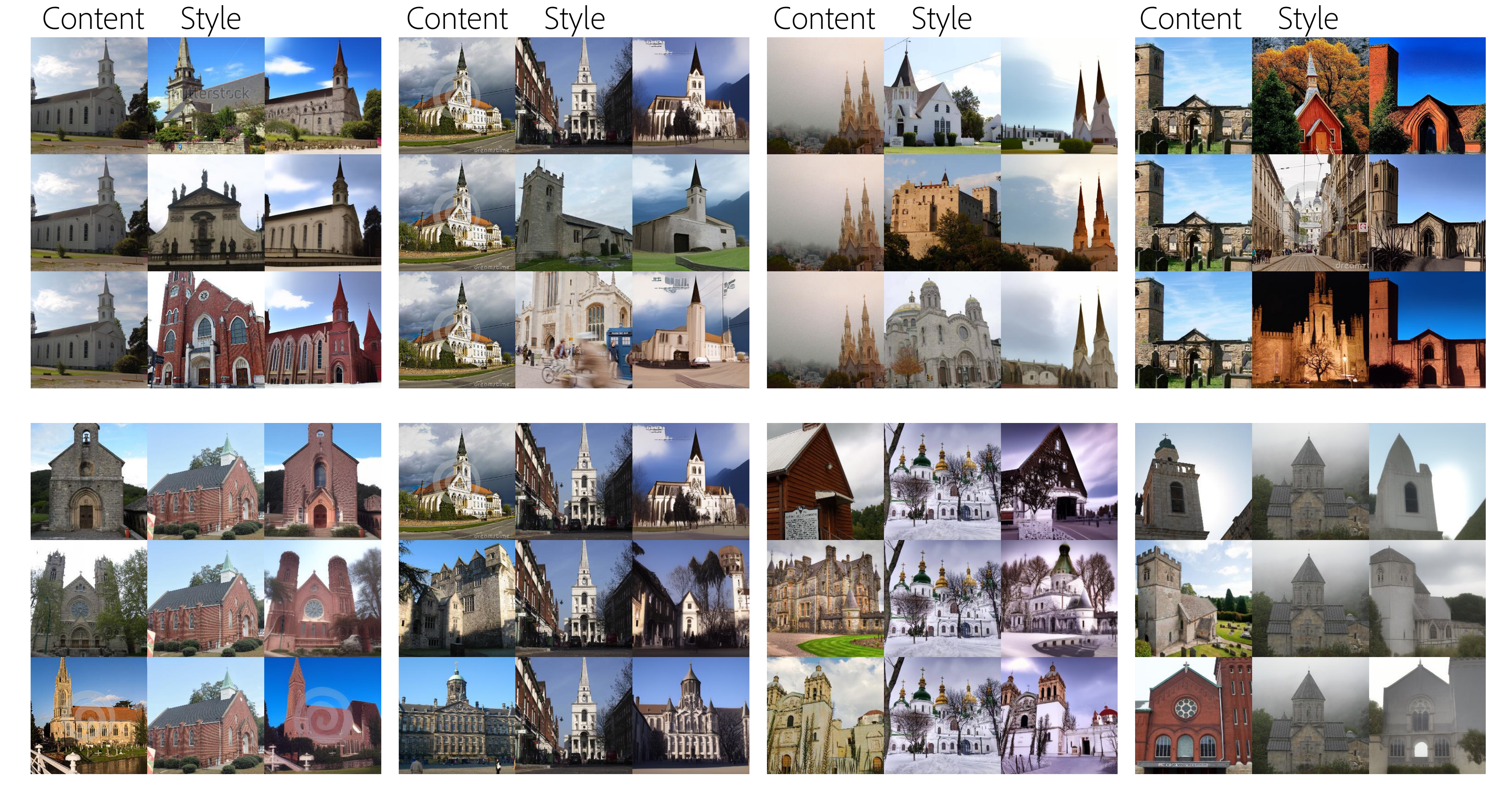}
\caption{Additional results on LSUN-church. The results are sampled by \emph{reverse DDIM}.}
\label{fig:LSUN churches aditonal results}
\end{figure*}

\begin{figure*}[h]
\includegraphics[width=\textwidth]{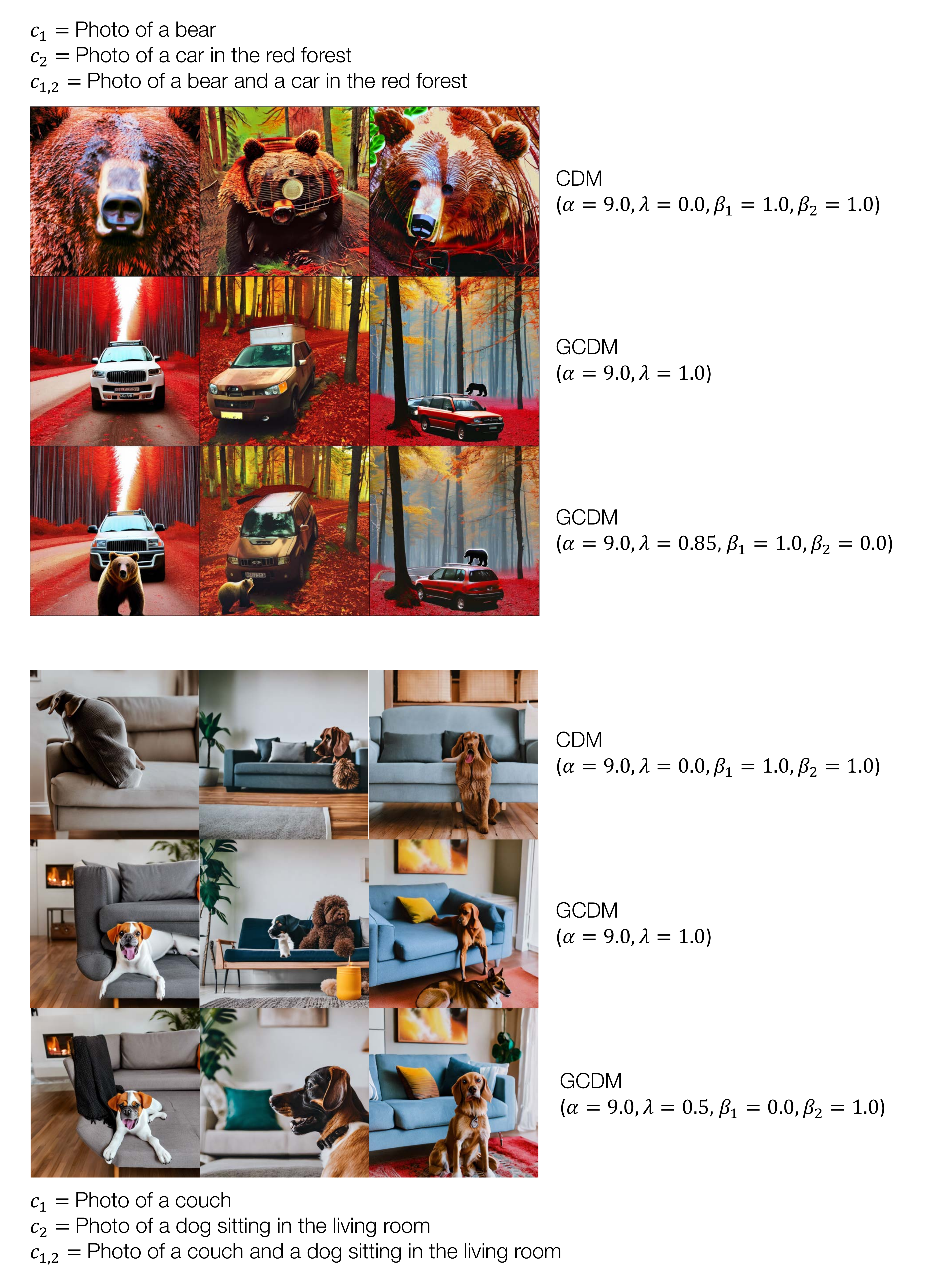}
\caption{Text2image synthesis results with GCDM hyperparameters.}
\label{fig:supp_text2image_hyperparameters}
\end{figure*}

Fig.~\ref{fig:LSUN_churches autoencoder} shows the effect of the starting point $x_T$ given same content and style codes.
Fig.~\ref{fig:FFHQ aditonal results} shows the additional results on FFHQ dataset. Fig.~\ref{fig:LSUN churches aditonal results} shows the results of multiple style and single content, and vice versa on LSUN church dataset. 
Fig.~\ref{fig:supp_text2image_hyperparameters} shows the hyperparameters used for CDM and GCDM on Stable Diffusion V2.

\clearpage

\end{document}